%% file: main.tex
\documentclass[conference]{IEEEtran}
\IEEEoverridecommandlockouts
\usepackage{amssymb}
\usepackage{amsfonts}
\usepackage{amsmath}
\usepackage{algorithm}
\usepackage[noend]{algorithmic}
\usepackage{hyperref}
\usepackage{multirow}
\usepackage[table]{xcolor}
\usepackage{multicol}
\usepackage{xspace}
\usepackage{caption}
\usepackage{subcaption}
\usepackage{tablefootnote}
\usepackage{amsthm}
\usepackage[table]{xcolor}
\usepackage{graphicx}
\usepackage{forest,array}

\usepackage[mathscr]{euscript}

\newcommand{\specialcell}[2][c]{%
\begin{tabular}[#1]{@{}c@{}}#2\end{tabular}
}

\newcommand{\norm}[1]{\left\lVert#1\right\rVert}

\usepackage{pifont}

\usepackage{lineno,hyperref}
\modulolinenumbers[5]

\def\BibTeX{{\rm B\kern-.05em{\sc i\kern-.025em b}\kern-.08em
    T\kern-.1667em\lower.7ex\hbox{E}\kern-.125emX}}
\begin{document}

\title{An Analysis of the Preferences of Distribution Indicators in Evolutionary Multi-Objective Optimization\\

\thanks{Identify applicable funding agency here. If none, delete this.}
}
\author{\IEEEauthorblockN{Anonymous Authors}}

\author{\IEEEauthorblockN{1\textsuperscript{st} Jes{\'u}s Guillermo Falc{\'o}n-Cardona}
\IEEEauthorblockA{\textit{School of Engineering and Sciences} \\
\textit{Tecnologico de Monterrey}\\
Monterrey, N.L., M{\'e}xico, 64849\\
jfalcon@tec.mx}
\and
\IEEEauthorblockN{2\textsuperscript{nd} Mahboubeh Nezhadmoghaddam}
\IEEEauthorblockA{\textit{School of Engineering and Sciences} \\
\textit{Tecnologico de Monterrey}\\
Monterrey, N.L., M{\'e}xico, 64849\\
A00838402@tec.mx}

\and
\IEEEauthorblockN{3\textsuperscript{rd} Emilio Bernal-Zubieta}
\IEEEauthorblockA{\textit{School of Engineering and Sciences} \\
\textit{Tecnologico de Monterrey}\\
Monterrey, N.L., M{\'e}xico, 64849\\
A01570751@tec.mx}
}

\maketitle

\begin{abstract}

The distribution of objective vectors in a Pareto Front Approximation (PFA) is crucial for representing the associated manifold accurately. Distribution Indicators (DIs) assess the distribution of a PFA numerically, utilizing concepts like distance calculation, Biodiversity, Entropy, Potential Energy, or Clustering. Despite the diversity of DIs, their strengths and weaknesses across assessment scenarios are not well-understood. This paper introduces a taxonomy for classifying DIs, followed by a preference analysis of nine DIs, each representing a category in the taxonomy. Experimental results, considering various PFAs under controlled scenarios (loss of coverage, loss of uniformity, pathological distributions), reveal that some DIs can be misleading and need cautious use. Additionally, DIs based on Biodiversity and Potential Energy show promise for PFA evaluation and comparison of Multi-Objective Evolutionary Algorithms.
\end{abstract}


\input{Sections/S1}
\input{Sections/S2}
\input{Sections/S3}
\input{Sections/S4}

\input{Sections/S5}

\bibliographystyle{ieeetr}
\bibliography{main.bbl}

\end{document}

%% file: Sections/S1.tex
\section{Introduction} 

For more than thirty years, researchers have designed Multi-Objective Evolutionary Algorithms (MOEAs) to tackle complex Multi-Objective Optimization Problems (MOPs)~\cite{Brockhoff23}. Due to the mutual conflict of objectives, the solution of a MOP is a set (possibly infinite) of decision vectors, denoted as \emph{Pareto Set}, whose image is the so-called \emph{Pareto Front}. The goal of a MOEA is to generate a discrete solution set that properly represents the entire manifold associated with the Pareto Front when no preferences are \emph{a priori} articulated. Each MOEA generates a Pareto Front Approximation (PFA) with specific properties of convergence, uniformity, and spread according to its design principles. A complex challenge is to judge what a good PFA is, in terms of convergence, uniformity, and spread, for a given MOP due to the vast diversity of MOEAs. 

In the early days of Evolutionary Multi-Objective Optimization (EMOO), visualization was crucial for evaluating PFAs embedded into two- and three-objective spaces. A visual comparison is subjective and lacks precision, even in these low-dimensional objective spaces. Moreover, a visual comparison is more difficult when tackling MOPs with more than three objectives\footnote{In EMOO, a MOP with more than three objectives is usually denoted as a Many-Objective Optimization Problem. However, for simplicity, we use the acronym MOP regardless of the number of objectives.}. To alleviate this issue, Quality indicators (QIs), which are set functions, quantitatively assess PFAs, allowing a more precise comparison of convergence, uniformity, and spread~\cite{Li19,Audet21}. QIs serve multiple purposes, such as comparing PFAs, monitoring their search process, defining stopping criteria, and guiding search as a selection mechanism~\cite{Li19}. In this paper, we focus on QIs for comparing PFAs.

Three key factors determine the quality of a PFA: convergence, uniformity, and spread~\cite{Zitzler00}. Convergence measures the proximity of the solutions to the Pareto Front, thus, making it a critical aspect to compare PFAs (and, therefore, MOEAs). Consequently, many efforts have been focused on analyzing convergence QIs, such as the hypervolume indicator (HV)~\cite{Li19,Audet21,Guerreiro21,Cai22}. On the other hand, uniformity and spread can be regarded as complementary characteristics that describe the evenness and the coverage of the points over the Pareto Front, respectively~\cite{Emmerich13a}. Consequently, uniformity and spread are crucial to evaluate how well a convergent PFA represents a Pareto Front. However, uniformity and spread QIs have received less attention from the EMOO community. 

According to Wang~\emph{et al.}, there is another characteristic denoted as diversity~\cite{Wang17}. A PFA has a good degree of diversity if its solutions cover a large portion of the range of each objective, i.e., they provide a decision maker enough information about the Pareto Front. In contrast, a uniform PFA has its solutions evenly distributed, failing to cover the whole range of each objective. At first glance, uniformity and diversity\footnote{Let $\mathcal{A}=\{\vec{a}_1, \dots, \vec{a}_N\}, \vec{a}_i \in \mathbb{R}^m$ be a PFA and $D$ be a point-to-set dissimilarity measure. Wang~\emph{et al.} defined $\text{Diversity}(\mathcal{A}) = \sum_{\vec{a} \in \mathcal{A}} D(\vec{a}, \mathcal{A} \setminus \{\vec{a}\})$ and $\text{Uniformity}(\mathcal{A})=\text{var}_{\vec{a} \in \mathcal{A}} D(\vec{a}, \mathcal{A} \setminus \{\vec{a}\})$, where $\text{var}(\cdot)$ calculates the variance~\cite{Wang17}.} seem dissimilar, but they are two versions of how solutions are distributed over the Pareto Front. In other words, uniformity and diversity are two specifications of a more general concept that we denote as \emph{distribution}. However, it is worth emphasizing that uniformity does not always imply good diversity, and vice versa. 

The utilization of Distribution Indicators (DIs) dates back to the mid-90s when Srinivas and Deb proposed the $\chi$-like Deviation Indicator to measure the distribution of low-dimensional PFAs. In 1995, Schott introduced the Spacing Indicator (SP) that measures the average distance deviation of the points in a PFA~\cite{Schott95}. For several years, SP was one of the preferred DIs, but some authors pointed out issues of SP when assessing PFAs, giving rise to modifications of SP~\cite{Bandy04,Meng05}. Other DIs based on distance calculations as SP have been proposed, for instance, the Uniformity Level (UNL)~\cite{Sayin00}, which in recent years has been employed to design subset selection algorithms~\cite{Shang21}. Aside from using distances, Shannon's Entropy has inspired the definition of a DI where, although counter-intuitive, the aim is to maximize its value~\cite{Mehr03}. Clustering techniques, such as $k$-Nearest Neighbors ($k$-NN) and hierarchical clustering, have also encouraged the construction of DIs~\cite{Wang12,Li05}. In recent years, more sophisticated DIs have been introduced. In 2017, Wang~\emph{et al.} defined the Pure Diversity Indicator (PUD), which measures the diversity of a PFA~\cite{Wang17}. It is worth noting that PUD is based on the Weitzman Diversity, which is a general definition of Biodiversity proposed in 1992~\cite{Weitzman92}. The Weitzman Diversity is also the baseline of the Solow-Polasky Diversity Indicator (SPD) that has been scarcely studied in EMOO~\cite{Emmerich13a,Solow94}. Other concepts used as the baseline of DIs are Potential Energy~\cite{Falcon23} and search directions~\cite{Cai18}, among others. Despite the different sources of inspiration to design DIs, there is no common consensus on a good DI, as in the case of HV for convergence QIs. Moreover, it lacks information about the strengths and weaknesses of state-of-the-art DIs when utilized under different assessment scenarios. 

In this paper, we first propose a taxonomy to classify the DIs proposed in the EMOO community. To the authors' best knowledge, this is the first taxonomy for this aim. Based on the proposed taxonomy, we study the strengths and weaknesses of nine DIs, which represent each taxonomic category. Concretely, we analyze: 1) the Diversity Indicator based on Reference Vectors (DIR)~\cite{Cai18}, 2) PUD~\cite{Wang17}, 3) SPD~\cite{Solow94}, 4) Riesz $s$-energy (RSE)~\cite{Sergiy19,Falcon23}, 5) Information-Theoretic Metric Entropy Indicator (ENI)~\cite{Mehr03}, 6) Coverage over the Pareto Front (CPF)~\cite{Tian19}, 7) $k$-NN based Uniformity Indicator (KUA)~\cite{Wang12}, 8) Diversity Metric based on Clustering (CDI)~\cite{Li05}, and 9) UNL~\cite{Sayin00}. We analyze each DI's behavior under loss of coverage, loss of uniformity, and PFAs with three types of pathological distributions. Moreover, we provide some recommendations for the utilization of the DI. This paper's ultimate goal is to highlight DIs' promising design strategies.

The remainder of this paper is as follows. Section~\ref{sec:background} provides the fundamental mathematical concepts of Multi-Objective Optimizations and QIs. Section~\ref{sec:DIs} describes Distribution QIs and some representative examples based on our proposed taxonomy. Section~\ref{sec:results} is devoted to presenting a numerical study of the properties of the selected Distribution Indicators. Finally, our main conclusions and future research work are sketched in Section~\ref{sec:conclusions}.

%% file: Sections/S2.tex
\section{Background}\label{sec:background} 
This section defines the basic mathematical terminology of Multi-Objective Optimization. Moreover, we formalize what a PFA and QIs are. 

Without loss of generality, an unconstrained MOP with $n$ decision variables and $m(\ge 2)$ conflicting objectives is defined as follows:
\begin{equation}
\min_{\vec{x} \in \Omega} \left \{f(\vec{x}) := [f_1(\vec{x}), f_2(\vec{x}), \ldots, f_m(\vec{x})]^\intercal \right \} 
\end{equation}

\noindent where $\vec{x}=[x_1,\dots,x_n]^\intercal$ is a decision vector and $\Omega \subseteq \mathbb{R}^n$ is the decision space. $f:\Omega \rightarrow \Lambda$ (where $\Lambda \subseteq \mathbb{R}^m$ is the objective space) is an objective vector of $m \ge 2$ objectives, $f_i :\Omega \rightarrow \mathbb{R}$ for $i=1,2,\dots, m$, mutually in conflict.

Due to the conflict among the objectives, the optimal solution of a MOP is a set of decision vectors. We employ the Pareto dominance relation ($\prec$) that induces a strict partial order in $\Omega$ to recognize these optimal decision vectors. Given $\vec{x} ,\vec{y} \in \Omega$, we say that $\vec{x}$ Pareto dominates $\vec{y}$ (denoted as $\vec{x} \prec \vec{y}$) if $f_i(\vec{x}) \le f_i(\vec{y})$ for all $i=1,2,\dots,m$, and there is at least an index $j \in \{1,2,\dots,m\}$ such that $f_j(\vec{x}) < f_j(\vec{y})$. A solution $\vec{x}^* \in \Omega$ is Pareto optimal if there is no other $\vec{x} \in \Omega$ such that $\vec{x} \prec \vec{x}^*$. The Pareto Set (PS) is the set of all Pareto optimal solutions, and its image in $\Lambda$ is the so-called Pareto Front, which is defined as $\text{PF} = \{ f(\vec{x}^*) \: | \: \vec{x}^* \in \text{PS} \}$. PF represents the best possible trade-offs among the objectives. Two special objective vectors enclose a Pareto Front. The first one is the ideal vector ($\vec{z}^* \in \mathbb{R}^m$) where $z_i^* = \min_{\vec{x} \in \Omega} f_i(\vec{x})$, $i=1,2,\dots,m$. The other one is the nadir vector ($\vec{z}^{\text{nad}}$) which is an anti-optimal point where $z_i^{\text{nad}} = \max_{\vec{x}^* \in \text{PS}} f_i(\vec{x}^*)$, $i=1,2,\dots,m$.

Since a PS may be infinite, a MOEA must generate a discrete representation of it. A Pareto Set Approximation (PSA) obtained by an MOEA can be defined as $\text{PSA} = \{\vec{x} \:|\: \vec{x} \not \prec \vec{y}  \wedge \vec{y} \not \prec \vec{x}, \vec{x}, \vec{y} \in \Omega \}$. In consequence, a Pareto Front Approximation (PFA) is the image of a PSA, i.e., $\text{PFA} = f(\text{PSA})$. Quality Indicators (QIs) were introduced to assess the convergence, distribution, and spread of PFAs numerically, thus, comparing MOEAs~\cite{Li19,Audet21}. A unary QI is a set function ($\mathcal{I}$) that assigns a real value to a PFA, depending on its specific preferences. In general, a $k$-ary QI takes as input $k$ PFAs and then evaluates their quality relative to each other to produce a single real value. In this paper, we focus on unary QIs that measure the Distribution of a PFA, i.e., Distribution Indicators. 

%% file: Sections/S3.tex
\section{Distribution Indicators}\label{sec:DIs} 

\input{Figs/Taxonomy/taxonomy_wide}

Currently, many DIs' design principles come from different fields~\cite{Li19,Audet21}. However, to the authors' best knowledge, no taxonomy completely classifies the distinct types of DIs. Consequently, we propose a taxonomy for DIs in Figure~\ref{fig:taxonomy}. To design the taxonomy, we examined papers published from 1992 to the end of 2023 that specifically consider the distribution in the objective space. It is worth emphasizing that most of the references were collected from the survey papers of Li~\emph{et al.}~\cite{Li19} and Audet~\emph{et al.}~\cite{Audet21}. Furthermore, we should mention that we contemplated DIs but not Spread QIs in designing the taxonomy. Based on the literature analysis, we classified DIs into four subclasses: 1) Neighborhood-based DIs, 2) Distance-based DIs, 3) Cluster-based DIs, and 4) Hybrid DIs. In turn, Neighborhood-based DIs are subdivided into a) Direction-based DIs, b) Biodiversity-based DIs, c) Potential Energy-based DIs, d) Entropy-based DIs, and e) Grid-based DIs. Hence, our taxonomy classifies DIs into eight classes that will be described in the next sections. Due to lack of space, we cannot provide the classification of all the DIs that have been published since 1992. Nevertheless, we summarize the characteristics of a representative DI per each class in Table~\ref{tab:DIs_table}. We should mention that the DIs appearing in Table~\ref{tab:DIs_table} are indicators that exclusively measure distribution, and they are the ones that we are studied in Section~\ref{sec:results}. We do not consider convergence QIs that can also be used to measure distribution, such as HV~\cite{Jiang16} for this paper. In the following, let $\mathcal{A} = \{\vec{a}_1,\dots,\vec{a}_N\}$, $\vec{a}_i \in \Lambda$ be a set of $N$ objective vectors that represent a PFA.

\input{Tables/DIs_table}

\subsection{Neighborhood-based DIs}
The DIs in this class measure the degree of variation between each point in a PFA and its surrounding neighborhood. This involves calculating functions such as Entropy, Potential Energy, and minimal distance between solutions. This category is divided into five subclasses since many DIs fall into it.

\subsubsection{Direction-based DIs}
The main characteristic of this subclass is the utilization of reference vectors to characterize distinct areas in the objective space. An example of this subclass is the Diversity Indicator based on Reference Vectors (DIR)~\cite{Cai18}. DIR partitions the objective space into subregions via a set $W=\{\vec{w}_1,\dots,\vec{w}_M\}$ of reference vectors\footnote{The authors propose the use of weight vectors. A vector $\vec{w} \in \mathbb{R}^m$ is a weight vector if $\sum_{i=1}^m w_i = 1$ and $w_i \ge 0$ for all $i=1,2,\dots,m$.}. Then, the points in the PFA are associated with a reference vector using an angular measure as follows:

\begin{equation}
\text{angle}(\vec{a}, \vec{w}) := \arccos\left(\frac{\vec{w}{}^\intercal \cdot (\vec{a} - \vec{z}^*)}{\lVert \vec{w} \rVert \cdot \lVert {\vec{a} - \vec{z}^*} \rVert}\right).
\end{equation}

\noindent The association of points in $\mathcal{A}$ with reference vectors defines a coverage vector $\vec{c}$ very similar to a niching measure. The value of DIR, which contemplates both spread and uniformity (according to the authors), is estimated based on the mean and standard deviation of the components of $\vec{c}$. The range of the indicator is $[0, \frac{M}{N}\sqrt{N-1}]$ and its value is to be minimized.

\subsubsection{Biodiversity-based DIs}
Influenced by Biology, this category focuses on measuring diversity in sets, particularly dissimilarity among elements. According to our literature review, only two DIs in EMOO are based on Biodiversity, namely, PUD~\cite{Wang12} and SPD~\cite{Solow94}. Both indicators are based on the Weitzman Diversity ($\mathcal{W}$), which is a recursive mathematical definition of Biodiversity that measures the dissimilarity between species~\cite{Weitzman92}. The Weitzman Diversity is defined as follows:

\begin{equation}\label{eq:weitzman}
\mathcal{W}(\mathcal{A}) := \max_{\vec{a} \in \mathcal{A}} \{\mathcal{W}(\mathcal{A} \setminus \{\vec{a}\}) + D(\vec{a},\mathcal{A} \setminus \{\vec{a}\})\},
\end{equation}

\noindent where $D(\vec{a}, \mathcal{A} \setminus \{\vec{a}\}) = \min_{\vec{u} \in \mathcal{A} \setminus \{\vec{a}\}} d\left(\vec{a}, \vec{u}\right)$ is a point-to-set measure of dissimilarity based on a distance measure\footnote{Weitzman indicated that is not mandatory that $d$ is a distance metric, thus, allowing the possibility of non-holding the triangle inequality.}. We can see that $\mathcal{W}$ is a basic dynamic programming equation. Thus, its solution is computationally expensive~\cite{Emmerich13a,Weitzman92}. In consequence, some simplifications of $\mathcal{W}$ have been proposed, such as SPD and PUD. 

SPD simplifies the $\mathcal{W}$ calculation by considering the first recursive step. Let $\mathbf{C}^{N \times N}$ be a full-rank matrix where $c_{ij} = e^{-\theta\cdot \norm{\vec{a}_i - \vec{a}_j}}$, $\vec{a}_i, \vec{a}_j \in \mathcal{A}$, $i,j=1,\dots,N$, and $\theta > 0$. Now, let $\mathbf{M} = \mathbf{C}^{-1}$. Hence, SPD is given by:

\begin{equation}
    SPD(\mathcal{A}) := \sum\limits_{i=1}^{N} \sum\limits_{j=1}^{N} m_{ij}.
\end{equation}

\noindent $\theta$ is a user-supplied parameter that controls how fast a PFA tends to $N$ when the distance increases. It is recommended to set $\theta=10$~\cite{Emmerich13a}. SPD is to be maximized, and its value lies in the $[1, N]$ range.

Another Biodiversity-based DI is PUD~\cite{Wang17} that follows Eq.~\eqref{eq:weitzman}. The authors propose using $L_p$-norm with $p=0.1$ to avoid misleading results when assessing PFAs embedded in objective spaces of over three dimensions. Furthermore, the authors proposed an approximation algorithm to calculate PUD. PUD will be maximized, and its value is in $[0, \infty)$.

\subsubsection{Potential Energy-based DIs}
This taxonomic category refers to DIs that utilize concepts of discrete Potential Energy to measure the distribution of a PFA. Hardin and Saff initially proposed the discrete Riesz $s$-energy (RSE) to generate even distributions of points over an $m$-dimensional sphere~\cite{Sergiy19}. In 2013, Bhattacharjee~\emph{et al.} used RSE to measure the uniformity of solutions of PFAs. Currently, RSE has promoted the design of algorithms to generate reference sets with good diversity~\cite{Falcon23}. Given a parameter $s > 0$, RSE is defined as follows:
\begin{equation}
\text{RSE}(A) := \sum\limits_{i \neq j} \lVert \Vec{a_i}-\Vec{a_j} \rVert^{-s}.
\end{equation}

\noindent where $\norm{\cdot}$ denotes the Euclidean distance. As long as $s$ is close to zero, RSE prefers solutions on the boundary of the Pareto Front. $s$ is usually set to $m-1$, where $m$ is the number of objectives~\cite{Shankar17,Falcon23}. RSE is to be minimized, and its value is in the range $(0, \infty)$. In high-dimensional objective spaces, it is possible that the value of RSE overflows. Hence, we recommend to take $\ln(\text{RSE}(\mathcal{A}))$. 

\subsubsection{Entropy-based DIs}
Inspired by Claude Shannon's work on Information Theory, Entropy-based DIs aim to measure how solutions are evenly distributed within the Pareto Front. Consequently, Farhang-Mehr and Azarm proposed the Information Theoretic Metric Entropy Indicator (ENI)~\cite{Mehr03}. To adapt the Shannon's Entropy, ENI divides the objective space using a grid, where the range of each objective is divided into $T \in \mathbb{N}$ portions\footnote{To ease of comprehension, we use $T$ for all the objectives. However, the authors defined a specific $T_i$ for each objective $i=1,2,\dots,m$.}. Then, it measures the influence of all the points in a PFA to the center of each $m$-dimensional cell of the grid (i.e., a density value). The authors propose the employment of a Gaussian function as an influence function. For an $m$-dimensional PFA, ENI is given as follows:

\begin{equation}\label{eq:ENI}
\text{ENI}(\mathcal{A}) := - \sum\limits_{i_1=1}^{T} \sum\limits_{i_2=1}^{T} \cdots \sum\limits_{i_m=1}^{T} \rho_{i_1i_2...i_m} \ln({\rho_{i_1i_2...i_m}}),
\end{equation}

\noindent where $\rho_{i_1i_2...i_m}$ is the normalized density value over the cell $(i_1i_2...i_m)$. According to the authors, the flatter the landscape of ENI is, the more uniform the PFA is. In consequence, although counter-intuitive, ENI is to be maximized, with its range being $[0, \infty)$. Despite Eq.~\eqref{eq:ENI} being for any value of $m$, the authors recommended projecting the points of a PFA (embedded into an objective space of three or more dimensions) to a two-dimensional objective space.

\subsubsection{Grid-based DIs} 
The DIs of this category use a grid that divides the objective space as the backbone of their calculation. Hence, they place the points in a PFA into the generated hyperboxes to count their contribution to the overall distribution. It is worth noting that even though ENI uses a grid, its core is the utilization of Shannon's Entropy. Hence, in essence, it is not a pure Grid-based DI. One representative example of a Grid-based DI is the Coverage Over the Pareto Front (CPF)~\cite{Tian19}. Since a Pareto Front is a manifold of at most $m-1$ dimensions embedded into an $m$-dimensional objective space, assessing the distribution of the points in this space leaves empty most part of the space. In consequence, CPF projects the points in a PFA to reduce the objective space of dimension $m-1$. The projection is based on a unit simplex plane where every point in a PFA is replaced by its closest reference point to eliminate the convergence contribution. CPF then employs a uniform design method for the hypercube mapping. The distribution score is determined by comparing the volumes of the hypercubes associated with the PFA to those generated by a reference set ($\mathcal{Z}$ of size $M$). The range of this DI is $[0, 1]$ and its value must be maximized.

\subsection{Distance-based DIs}
The DIs under this class only employ distance calculations as their main mechanism to assess the distribution of a PFA. For instance, SP is a Distance-based DI~\cite{Schott95}. However, we do not consider it a representative indicator of this class because some authors pointed out that SP produces misleading results under specific scenarios~\cite{Bandy04,Meng05}. In consequence, we selected the Uniformity Level (UNL)~\cite{Sayin00} as the representative indicator of this taxonomic category because, in recent years, it has encouraged the generation of subset selection algorithms~\cite{Shang21}. Given a distance metric $d:\mathbb{R} \times \mathbb{R}^m \rightarrow \mathbb{R}$, UNL is defined as follows:

\begin{equation}
    \text{UNL}(\mathcal{A}) = \min\limits_{\substack{\vec{a}_i, \vec{a}_j \in A \\\vec{a}_i \neq \vec{a}_j}} d(\vec{a}_i,\vec{a}_j).
\end{equation}

\noindent The authors recommend setting $d$ as the Chebyshev distance. The range of this indicator is $[0, \infty)$, and its value is to be minimized.

\subsection{Cluster-based DIs}
Clustering techniques are well-known unsupervised learning methods in Machine Learning that aim to group similar objects. Hence, they have been adapted to measure the distribution of PFAs. The Diversity Metric based on Clustering (CDI) utilizes a hierarchical clustering to determine the ratio of resulting clusters to the cardinality of a PFA~\cite{Li05}. In the beginning, each point in the PFA is first considered as an individual cluster. Then, CDI identifies the shortest distance between any two clusters' centers. The clusters are combined if this distance is less than a user-supplied parameter $\overline{d}>0$. Otherwise, the algorithm terminates and $\text{CDI}(\mathcal{A}) = |\text{Clusters}(\mathcal{A})| / |\mathcal{A}|$. According to the authors, a too-small value of $\overline{d}$ may cause $\text{CDI}(\mathcal{A}) = 1$, which is the maximum value of the indicator. In contrast, a large $\overline{d}$ value leads to a CDI value close to zero. 

\subsection{Hybrid DIs}
This class of DIs is introduced because some indicators may share characteristics of the other categories, making its categorization difficult. As a representative example, Wang~\emph{et al.} proposed the $k$-NN based Uniformity Indicator (KUA)~\cite{Wang12}. KUA is a distance- and neighborhood-based DI. It employs an improved $k$-Nearest Neighbors algorithm with a clip strategy to enhance neighborhood size adaptability, ensuring a more reasonable evaluation. Based on the clustering, KUA measures the distribution in each cluster, returning a real value per cluster. To consider KUA as a unary DI, the authors proposed obtaining the mean distribution value of the clusters as the final value of KUA. The range of this indicator is $[0,1]$, and its value is to be minimized.

%% file: Figs/Taxonomy/taxonomy_wide.tex
\newcolumntype{C}[1]{>{\centering}p{#1}}
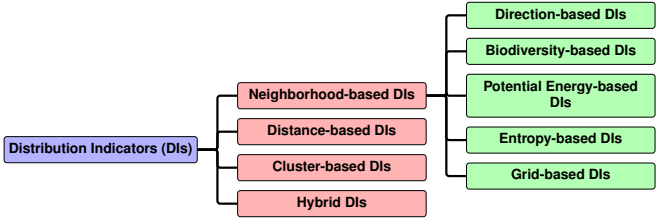
\begin{figure}
\resizebox{\linewidth}{!}{
\begin{centering}
\begin{forest}
for tree={
  if level=0{align=center}{
    align={@{}C{45mm}@{}},
  },
  grow=east,
  draw,
  font=\sffamily\bfseries,
  edge path={
    \noexpand\path [draw, \forestoption{edge}] (!u.parent anchor) -- +(5mm,0) |- (.child anchor)\forestoption{edge label};
  },
  parent anchor=east,
  child anchor=west,
  l sep=10mm,
  tier/.wrap pgfmath arg={tier #1}{level()},
  edge={ultra thick, rounded corners=2pt},
  if level=0{fill=blue!30}{if level=1{fill=red!30}{if level=2{fill=green!30}{}}},
  rounded corners=2pt
}
[Distribution Indicators (DIs)
  [Hybrid DIs]
  [Cluster-based DIs]
  [Distance-based DIs]
  [Neighborhood-based DIs
    [Grid-based DIs]
    [Entropy-based DIs]
    [Potential Energy-based DIs]
    [Biodiversity-based DIs]
    [Direction-based DIs]
  ]
]
\end{forest}
\end{centering}
}
\caption{Our proposed taxonomy for DIs.}\label{fig:taxonomy}
\end{figure}

%% file: Tables/DIs_table.tex
\newcommand{\trow}[7]{#1 & #2 & #3 & #4 & #5 & #6 & #7}

\begin{table*}[htp]
-\setlength\extrarowheight{3pt}
\centering
\scriptsize
\caption{Main properties of the analyzed DIs. Each DI is a representative of a given class in the proposed taxonomy. $N=|\mathcal{A}|$, $M=|W|=|\mathcal{Z}|$, and $m$ is the number of objectives.}
\label{tab:DIs_table}
\begin{tabular}{|c|c|c|c|c|c|c|}
\hline
\textbf{DI} & \textbf{\specialcell{Taxonomic \\ class}} & \textbf{Parameters} & \textbf{Complexity} & \textbf{Range} & \textbf{\specialcell{Maximize or \\ minimize}} & \textbf{Ref.} \\
\hline
\hline
\trow{DIR}{Direction-based}{$W \in \mathbb{R}^{M\times m}$}{$\Theta(mNM)$}{$[0,\frac{M}{N}\sqrt{N-1}$]}{Min}{\cite{Cai18}} \\
\trow{PUD}{Biodiversity}{$d:\mathbb{R}^m \times \mathbb{R}^m \rightarrow \mathbb{R}$}{$\Omega(mN^{3})$}{$[0, \infty)$}{Max}{\cite{Wang17}}\\
\trow{SPD}{Biodiversity}{$\theta > 0$}{$\Theta(mN^{3})$}{$[1, N]$}{Max}{\cite{Solow94}} \\
\trow{RSE}{Potential Energy-based}{$s > 0$}{$\Theta(mN^{2})$}{$(0,\infty)$}{Min}{\cite{Sergiy19}}\\
\trow{ENI}{Entropy-based}{$T \in \mathbb{N}$}{$\Theta(mN^2)$ or $\Theta(mN^m)$}{$[0, \infty)$}{Max}{\cite{Mehr03}} \\
\trow{CPF}{Grid-based}{$\mathcal{Z} \in \mathbb{R}^{M \times m}$}{$\Theta(mNM)$}{$[0,1]$}{Max}{\cite{Tian19}}\\
\trow{UNL}{Distance-based}{$d:\mathbb{R}^m \times \mathbb{R}^m \rightarrow \mathbb{R}$}{$\Theta(mN^{2})$}{$[0, \infty)$}{Min}{\cite{Sayin00}} \\
\trow{CDI}{Cluster-based}{$\overline{d} > 0$}{$\Theta(mN^{2})$}{$[0,1]$}{Max}{\cite{Li05}} \\
\trow{KUA}{Hybrid}{$k \in \mathbb{N}$}{$\Omega(mN^{2})$}{$[0,1]$}{Min}{\cite{Wang12}} \\
\hline
\end{tabular}

\end{table*}

%% file: Sections/S4.tex
\section{Experimental Results}\label{sec:results}
In this section, we analyze the preferences of DIR\footnote{We used the implementation available at the website of the author (\url{https://xinyecai.github.io/}).}, PUD\footnote{We used the PUD implementation available in PlatEMO.}, SPD, RSE, ENI, CPF\footnote{We used the CPF implementation available in PlatEMO.}, KUA, CDI, and UNL under three scenarios: loss of coverage, uniformity, and pathological distributions. The analysis of preferences is based on the assessment of PFAs that contain Pareto optimal solutions of MOPs in the Deb-Thiele-Laumanns-Zitzler (DTLZ)~\cite{Deb05c}, Walking-Fish-Group (WFG)~\cite{Huband06}, and the recently proposed Zapotecas-Coello-Aguirre-Tanaka (ZCAT)~\cite{Zapotecas23} test suites. Specifically, we selected the problems DTLZ1, DTLZ2, WFG1, WFG2, ZCAT1 - ZCAT4 with two to ten objectives because all of them exhibit different Pareto Front geometries but correlated with the shape of an $m$-dimensional simplex. The Pareto optimal solutions of the DTLZ and WFG problems were obtained from the PlatEMO platform~\cite{Tian17}, and the ones of the ZCAT suite were generated according to the algorithm described in~\cite{Zapotecas23}. All the initial PFAs contain approximately 10,000 normalized points to avoid biases due to the distinct scales of the objectives. To analyze the preferences of the selected DIs, we generated a high density of numerical values. However, due to lack of space, all these numerical results are presented in the Supplementary Material available at \url{http://tinyurl.com/dis-emoo}.

As indicated in Table~\ref{tab:DIs_table}, all the selected DIs require user-supplied parameters. We set these parameters as follows. We generated the set $W$ for DIR using the Two-Layered Simplex-Lattice Design (TLSLD) with cardinalities as specified in Table~\ref{tab:conf_coverage}. The $L_p$-norm with $p=0.1$ is used in PUD. We set $\theta=10$ for SPD. Regarding RSE, we set $s=m-1$ and we present it in logarithmic scale. For all the specifications of $m$, we set $T=100$ in ENI. We sampled $N$ (according to Table~\ref{tab:conf_coverage}) uniformly distributed points to generate the reference set for CPF, using the initial 10,000 points for each MOP. The Chebyshev function is used to calculate UNL. $\overline{d}$ is calculated for each instance, taking the maximum of the minimum distance of every PFA into consideration. Finally, $k=3$ for KUA.

\subsection{Loss of Coverage}

\input{Tables/Cardinality_coverage}

\begin{figure*}
    \centering
    \begin{subfigure}{0.18\textwidth}
        \includegraphics[width=\textwidth]{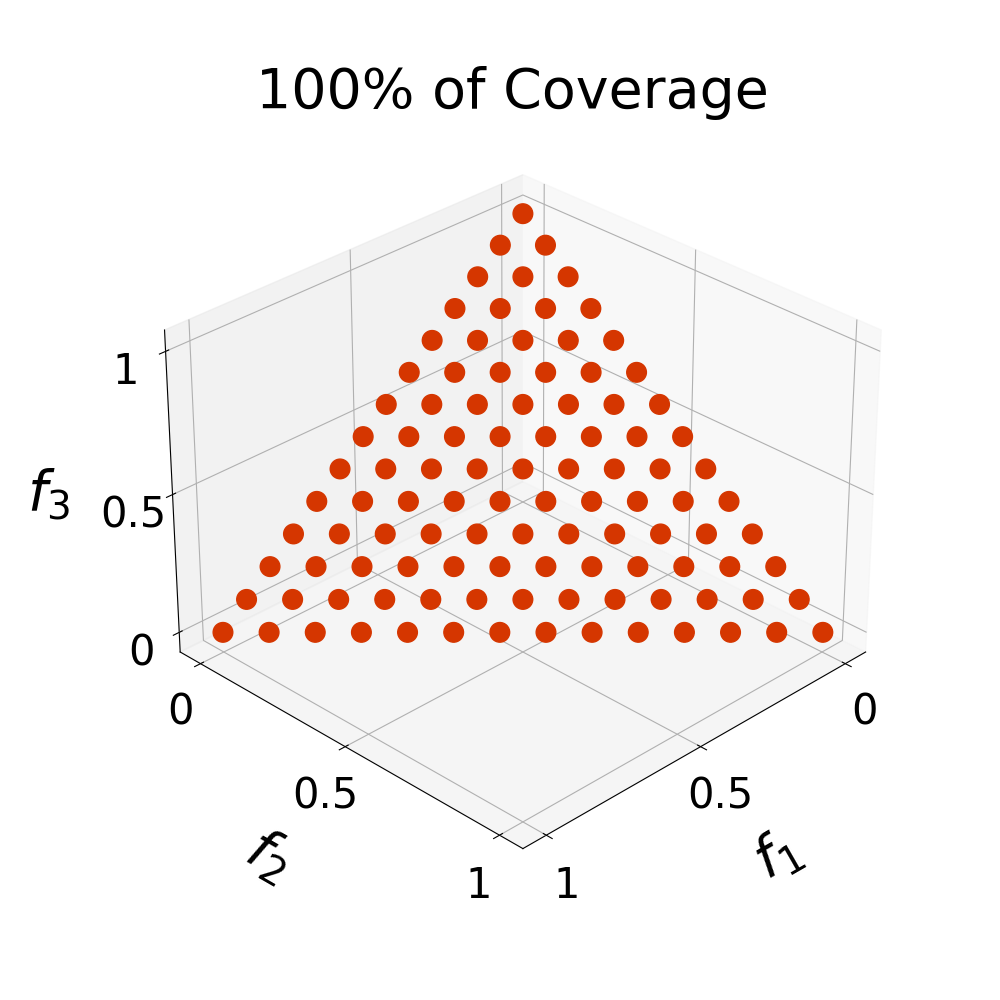}        
    \end{subfigure}
    \hspace{0.2cm}
      \begin{subfigure}{0.18\textwidth}
        \includegraphics[width=\textwidth]{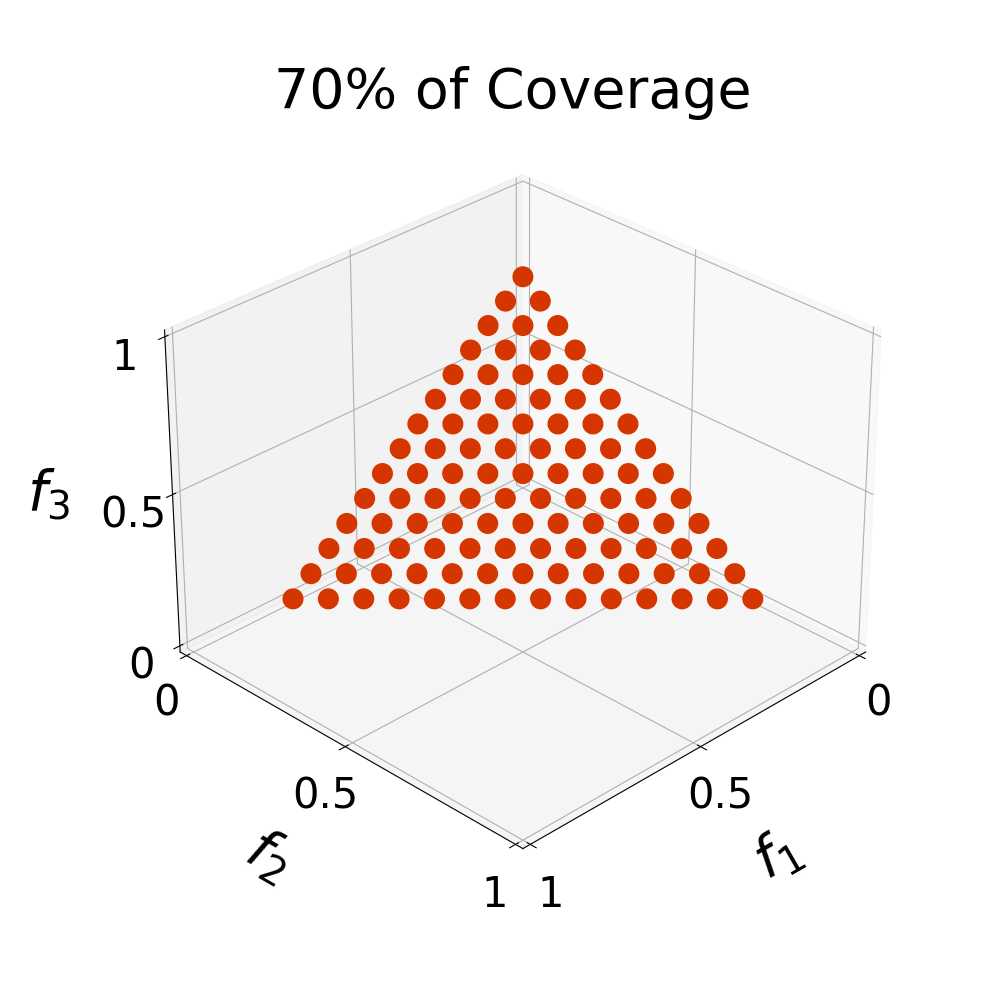}        
    \end{subfigure}
    \hspace{0.2cm}
    \begin{subfigure}{0.18\textwidth}
        \includegraphics[width=\textwidth]{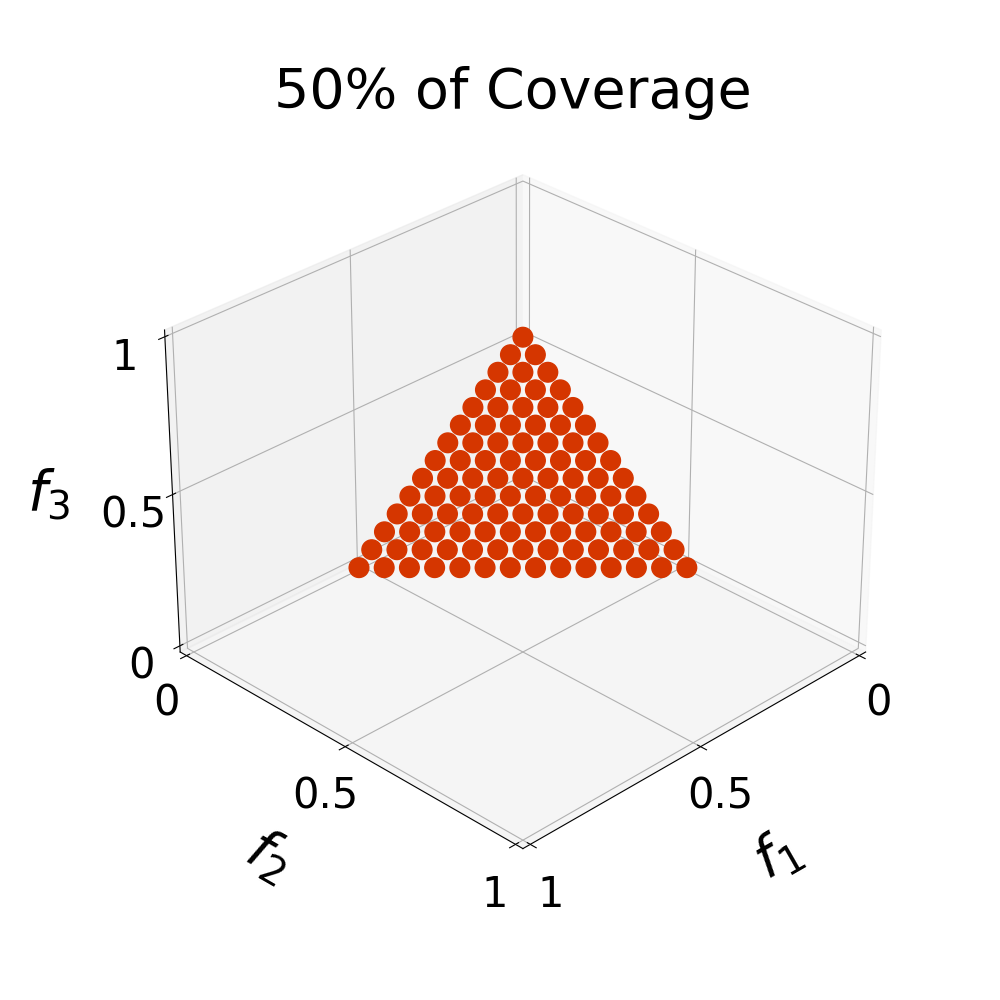}        
    \end{subfigure}
    \hspace{0.2cm}
      \begin{subfigure}{0.18\textwidth}
        \includegraphics[width=\textwidth]{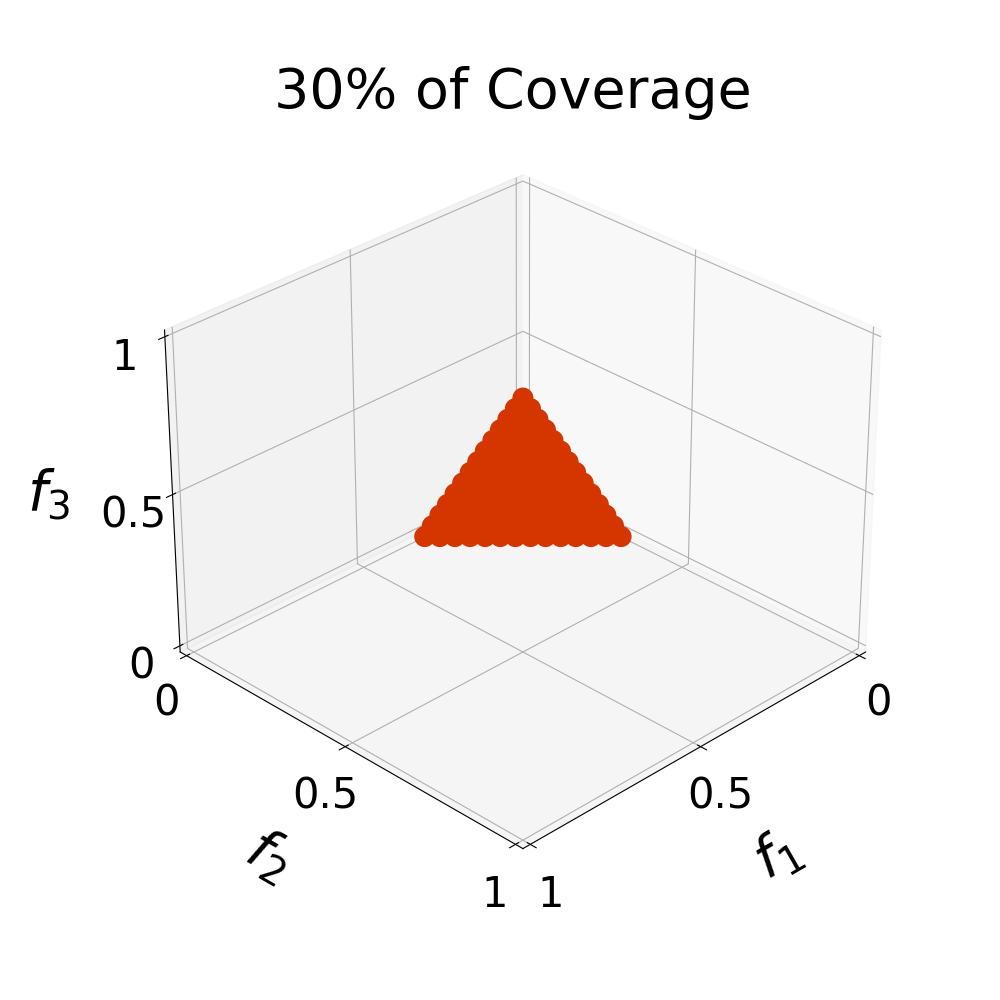}        
    \end{subfigure}
    \hspace{0.2cm}
    \begin{subfigure}{0.18\textwidth}
        \includegraphics[width=\textwidth]{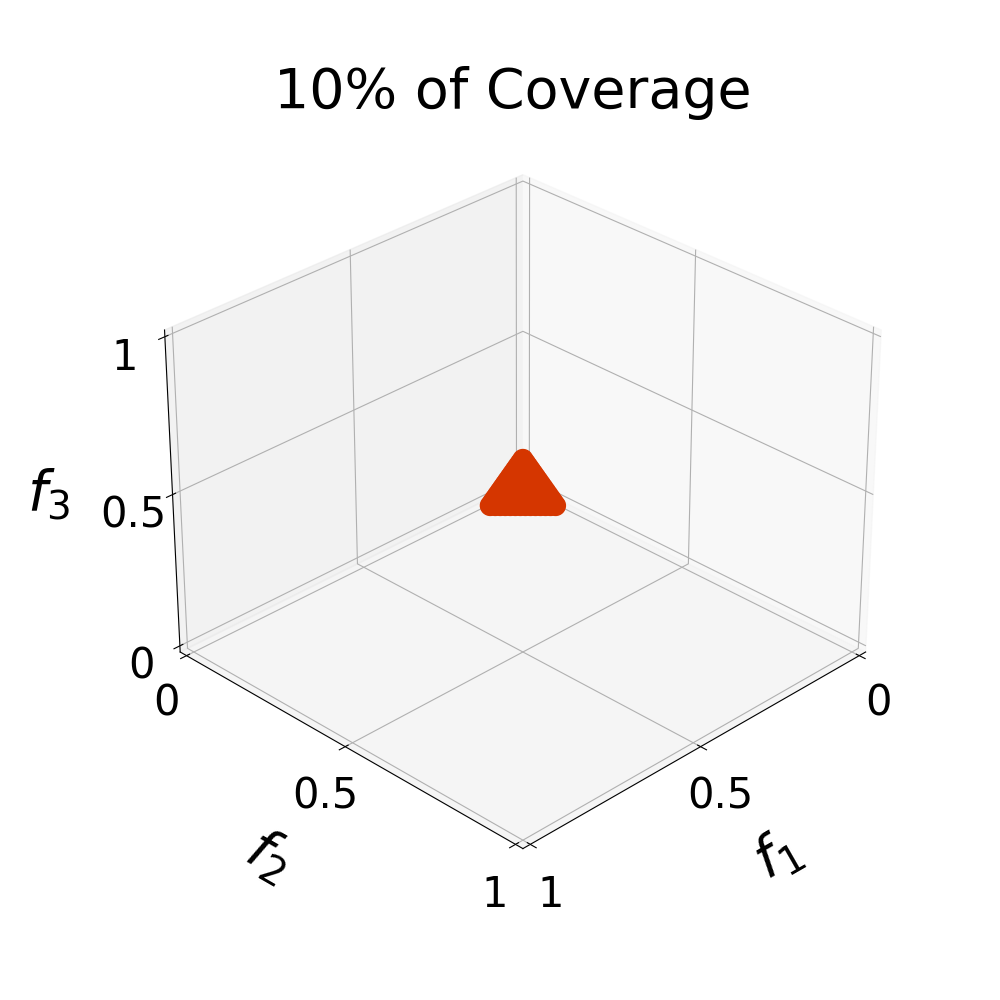}       
    \end{subfigure}

    \caption{Three-objective PFAs $\mathcal{L}_\gamma$ with $\gamma = 0.1,  0.5, 1.0$, simulating a loss of coverage. All the PFAs have 105 points.}\label{fig:coverage}
\end{figure*}

This experiment aims to analyze the preferences of the selected DIs when assessing PFAs with a different degree of coverage. We generated two types of PFAs (linear triangular and inverted linear triangular) given by $\mathcal{L} = \sum_{i=1}^m f_i - 1 = 0$ and $\mathcal{I} = \frac{1}{m-1}\sum_{i=1}^m f_i - 1 = 0$. We employed TLSLD to generate $N$ points (according to Table~\ref{tab:conf_coverage}) over $\mathcal{L}$ and $\mathcal{I}$. To simulate the loss of coverage, we shrinked $\mathcal{L}$ by a factor $\gamma \in \{0.1, 0.2, \dots, 1.0\}$ and, then, the points $\gamma \mathcal{L}$ were projected back to the hyperplane $\mathcal{L}$ to define the PFA $\mathcal{L}_\gamma$. A similar process was performed for $\mathcal{I}$. The ground truth for this experiment is $\mathcal{L}_{\gamma=1.0}$ and $\mathcal{I}_{\gamma=1.0}$ because they represent a PFA with 100\% of coverage as shown in Fig.~\ref{fig:coverage}. Consequently, the ideal behavior is that each DI prefers the ground truth over the PFAs with lower coverage. 

We evaluated all the instances with the selected DIs. The complete numerical results are shown in Tables SM-1 to SM-9 of the Supplementary Material. Figure~\ref{fig:coverage_preferences} summarizes the preferences of the DIs using a Likert Plot. We use a grading scale from 1 to 10 points for a PFA with $\gamma=0.1$ to $\gamma=1.0$, respectively. The numbers inside the boxes represent the average points given to the specific coverage case. Figure~\ref{fig:coverage_preferences} shows that both SPD and RSE perfectly grade all the PFAs; that is, they always prefer a PFA with a greater degree of coverage. This is a desired behavior since, when comparing two MOEAs, one may produce a PFA with lower coverage, and we do not want a DI to prefer this PFA. On the other hand, ENI and UNL exhibit a perfect inverse order in their preferences, which is not desirable, as we mentioned. DIR, PUD, CPF, and CDI prefer PFAs with greater coverage, but in some cases, they cannot recognize the ground truth. Therefore, they do not show perfect scores in Figure~\ref{fig:coverage}. 

A reason to use $\mathcal{L}$ and $\mathcal{I}$ is to analyze if the DIs are invariant to rotations of the PFA. Since $\mathcal{I}$ is an inverted $\mathcal{L}$, we want to verify if the indicator values change due to the rotation. Tables SM-1 to SM-9 in the Supplementary Material indicate that all the selected DIs are invariant to rotations. This is because the same indicator values are produced for all the specifications of $\mathcal{L}_\gamma$ and $\mathcal{I}_\gamma$.

\begin{figure}
    \centering
    \includegraphics[width=\columnwidth]{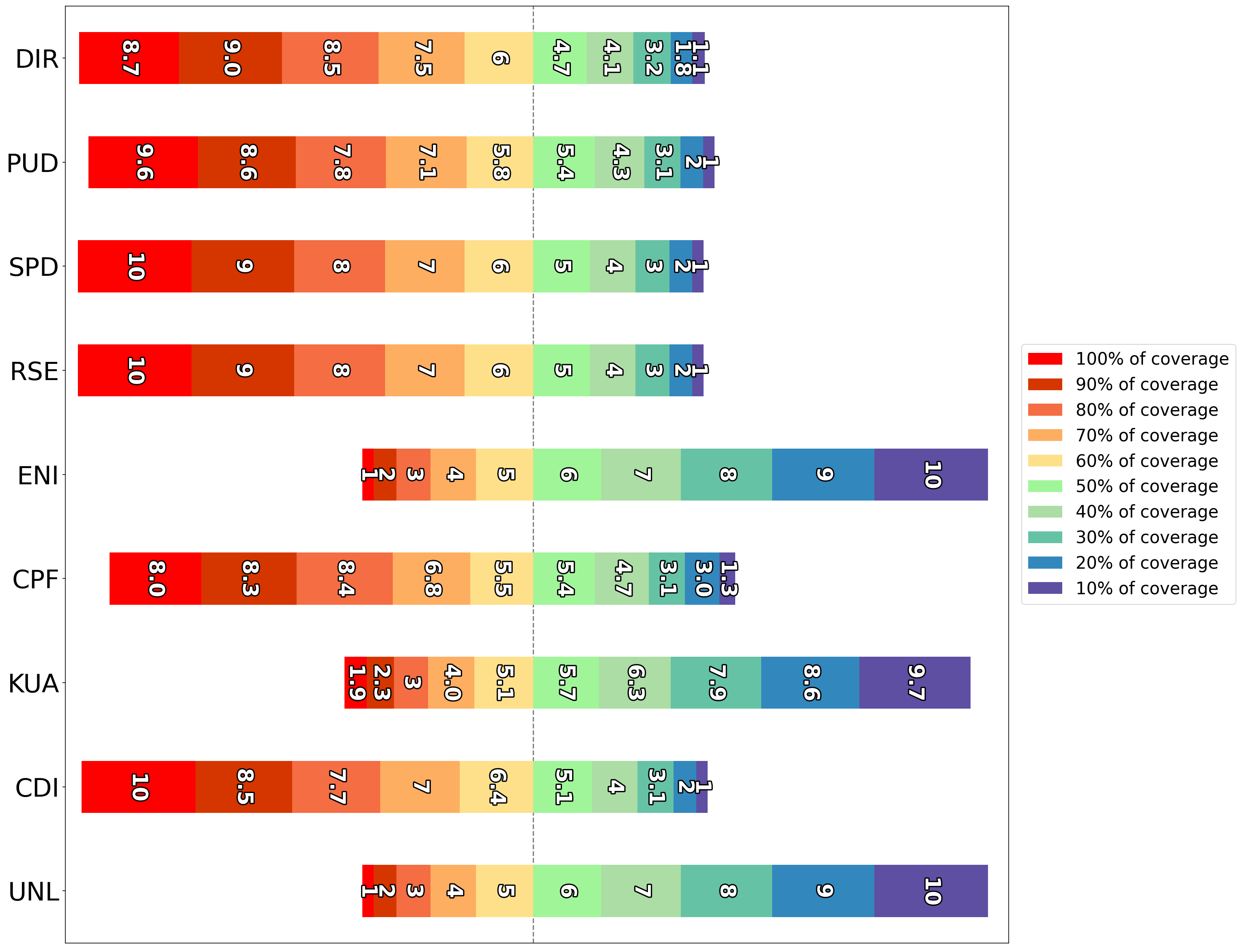}
    \caption{Preferences of the nine DIs for PFAs with different degrees of Coverage.}
    \label{fig:coverage_preferences}
\end{figure}

\subsection{Loss of Uniformity}

\begin{figure*}
    \centering
    \begin{subfigure}{0.18\textwidth}
        \includegraphics[width=\textwidth]{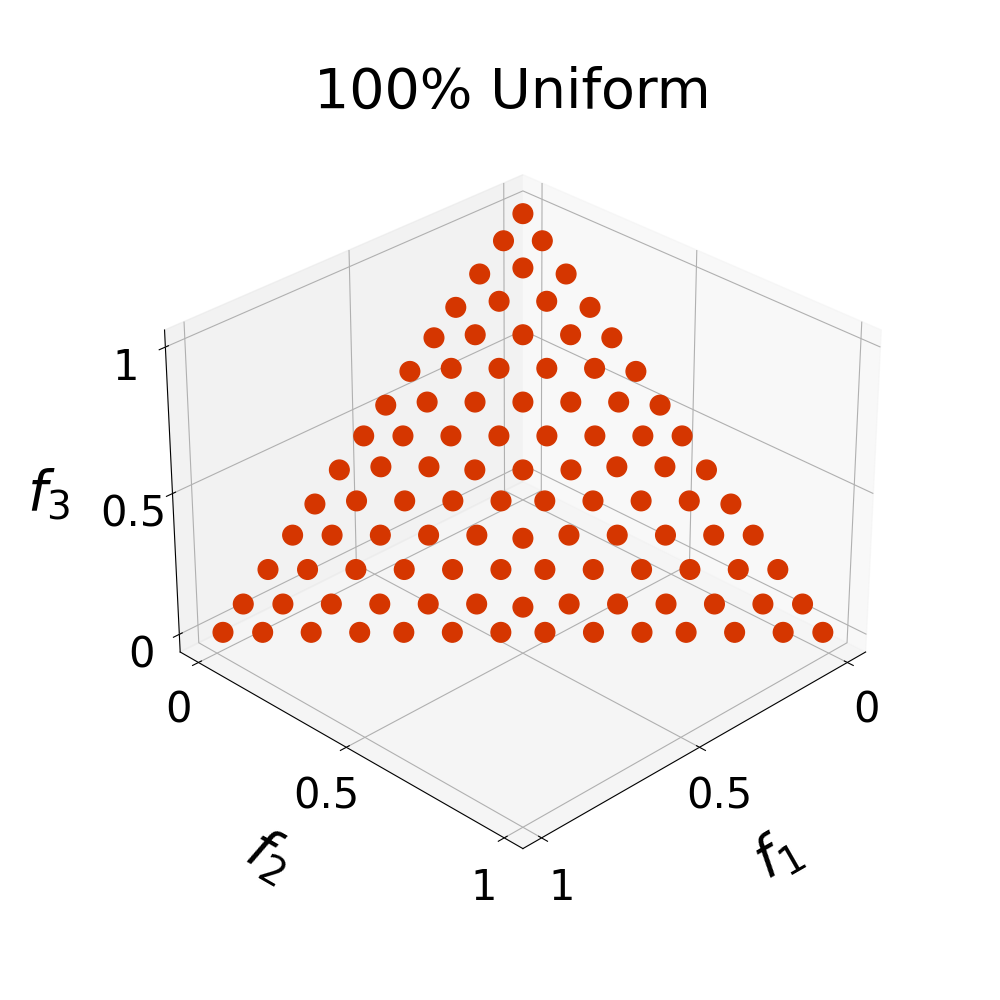}        
    \end{subfigure}
    \hspace{0.2cm}
      \begin{subfigure}{0.18\textwidth}
        \includegraphics[width=\textwidth]{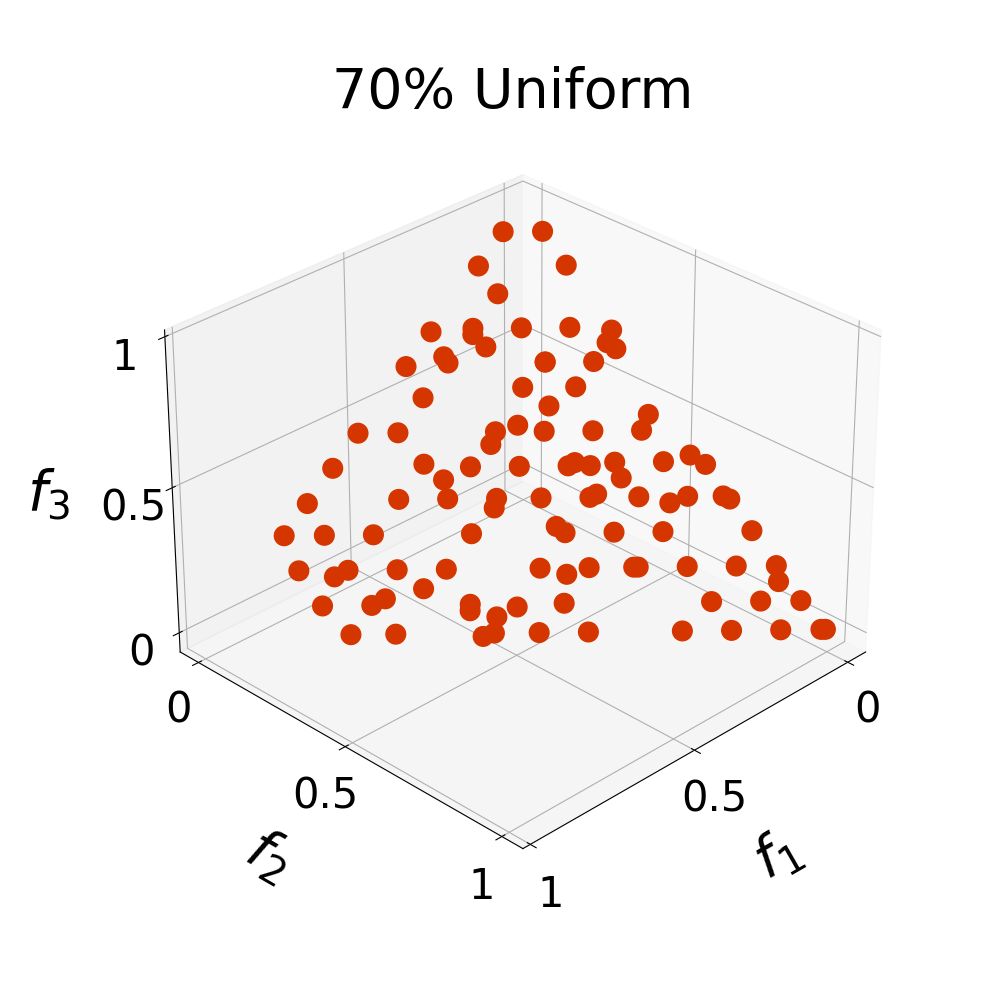}        
    \end{subfigure}
    \hspace{0.2cm}
    \begin{subfigure}{0.18\textwidth}
        \includegraphics[width=\textwidth]{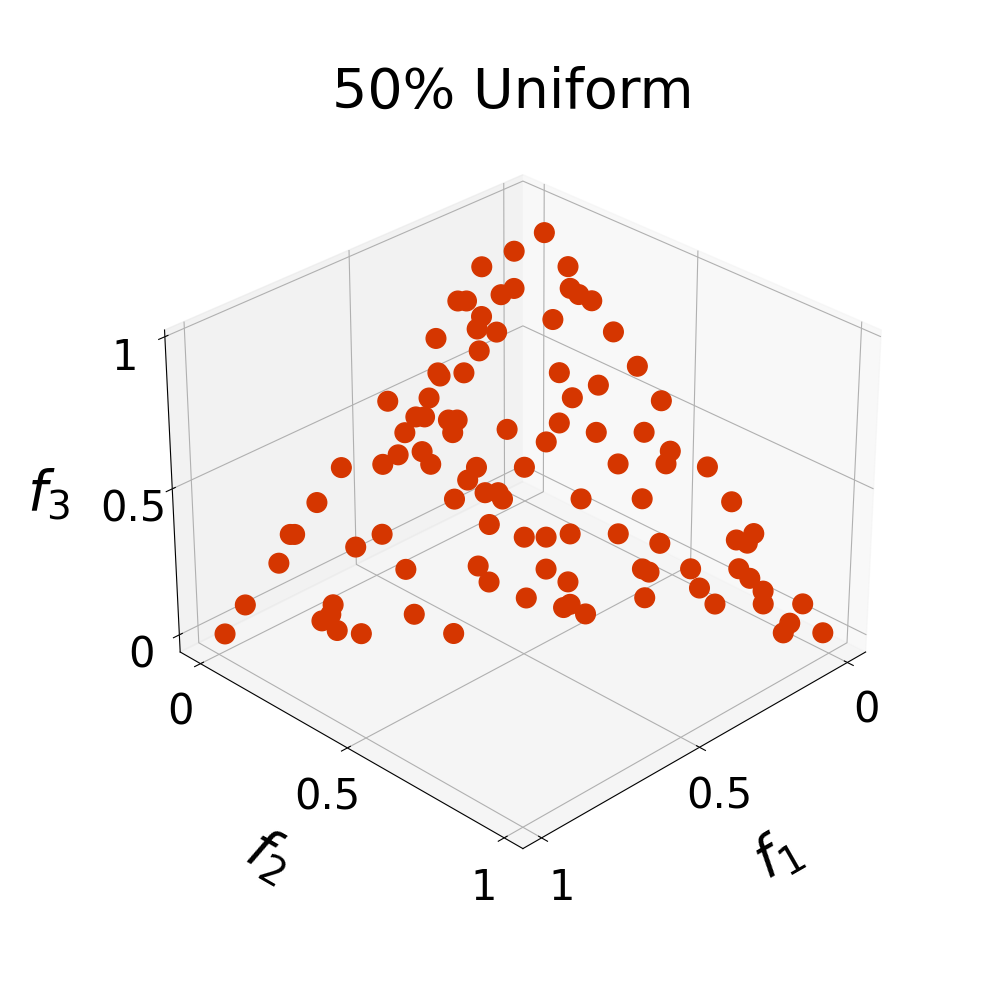}        
    \end{subfigure}
    \hspace{0.2cm}
      \begin{subfigure}{0.18\textwidth}
        \includegraphics[width=\textwidth]{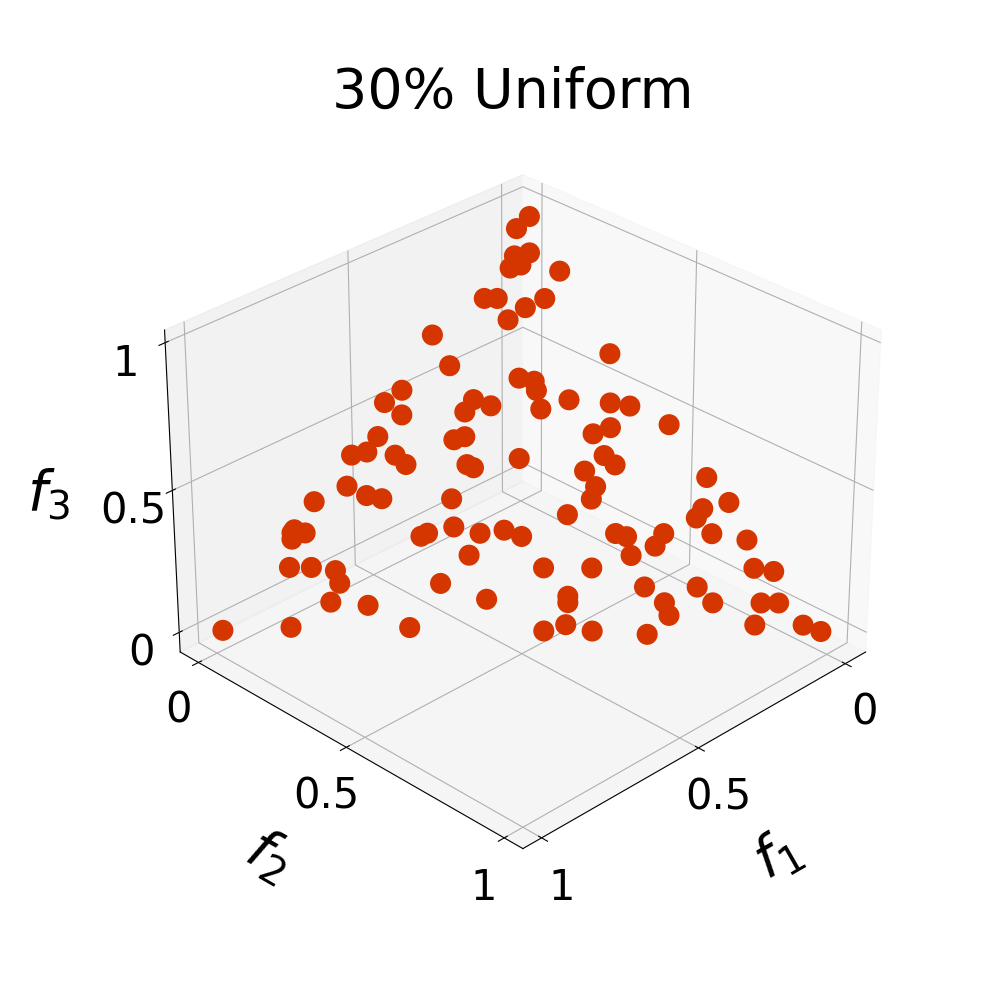}        
    \end{subfigure}
    \hspace{0.2cm}
    \begin{subfigure}{0.18\textwidth}
        \includegraphics[width=\textwidth]{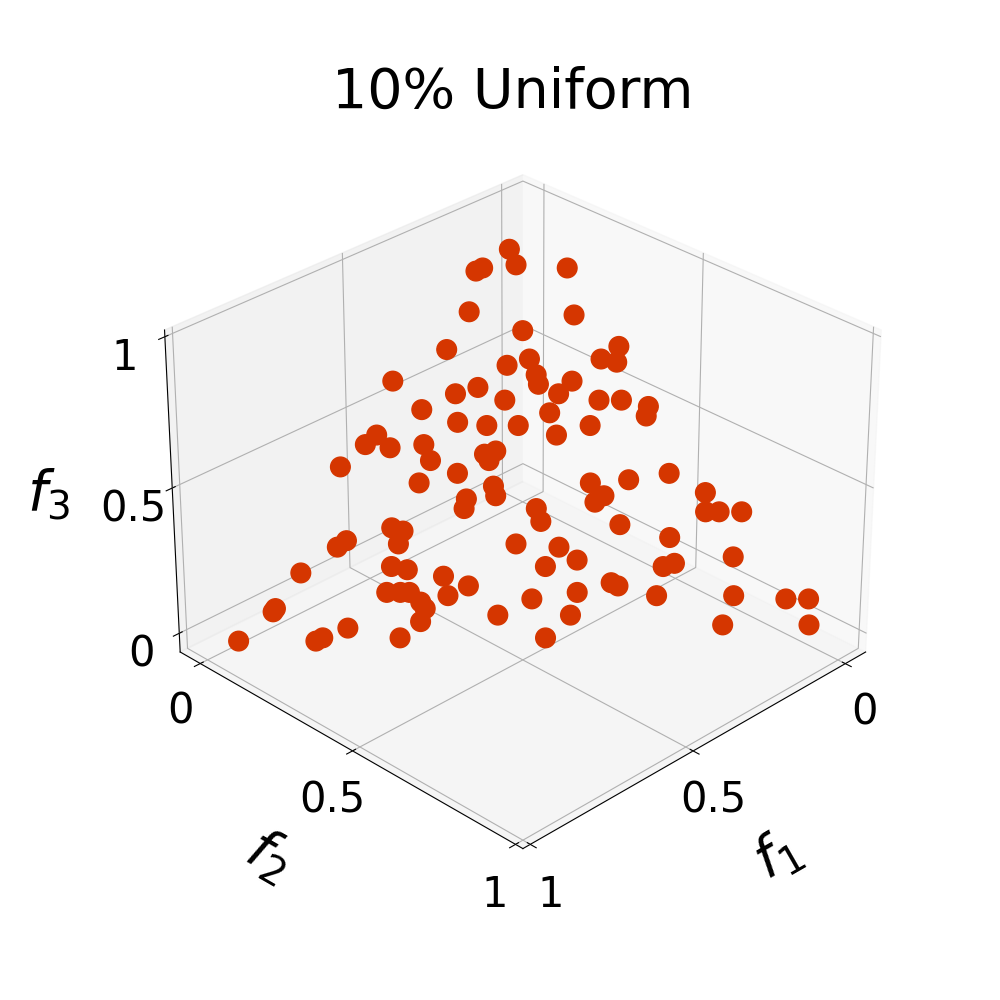}       
    \end{subfigure}
    

    \caption{Three-objective PFAs of  DTLZ1 for different degrees of Uniformity. All the PFAs have 105 points.}\label{fig:uniformity}
\end{figure*}

Now, we pay attention to the preferences of DIs when evaluating PFAs (we use DTLZ1, DTLZ2, WFG1, WFG2, and ZCAT1 - ZCAT4) with a different degree of uniformity. As we indicated at the beginning of Section~\ref{sec:results}, we obtained a high-density PFA ($\mathcal{S}$) of approximately 10,000 points for each MOP. For each specification of $m$, we obtained a subset ($\mathcal{A}$) of $N$ uniformly distributed solutions (according to Table~\ref{tab:conf_coverage}), where each $\vec{a} \in \mathcal{A}$ minimizes an achievement scalarizing function\footnote{Given $\vec{a} \in \mathbb{R}^m$, a reference point $\vec{z}\in \mathbb{R}^m$, and a weight vector $\vec{w}\in \mathbb{R}^m$, the achievement scalarizing function (ASF) is given by $u_{\vec{w}}^\text{ASF} (\vec{a}, \vec{z}) = \max_{i=1,2,\dots,m} \{w_i|a_i-z_i|\}$.} for a given weight vector ($\vec{w}$) generated by TLSLD. The set $\mathcal{A}$ is our ground truth since it is 100\% uniform, as illustrated in Figure~\ref{fig:uniformity}. To simulate the loss of Uniformity, we defined sets $\mathcal{A}_\beta$, where $\beta\in\{10,20,\dots, 90\}$, represent a PFA with a $\beta$\% of uniformity. To generate a $\mathcal{A}_\beta$, we randomly selected $\beta$\% of points from $\mathcal{A}$ and the remaining points are randomly obtained from $\mathcal{S}$, avoiding duplicity. Figure~\ref{fig:uniformity} compares PFAs for the three-objective DTLZ1 with $\beta = 100, 70, 50, 30,$ and 10 percent of uniformity.

\begin{figure}
    \centering
    \includegraphics[width=\columnwidth]{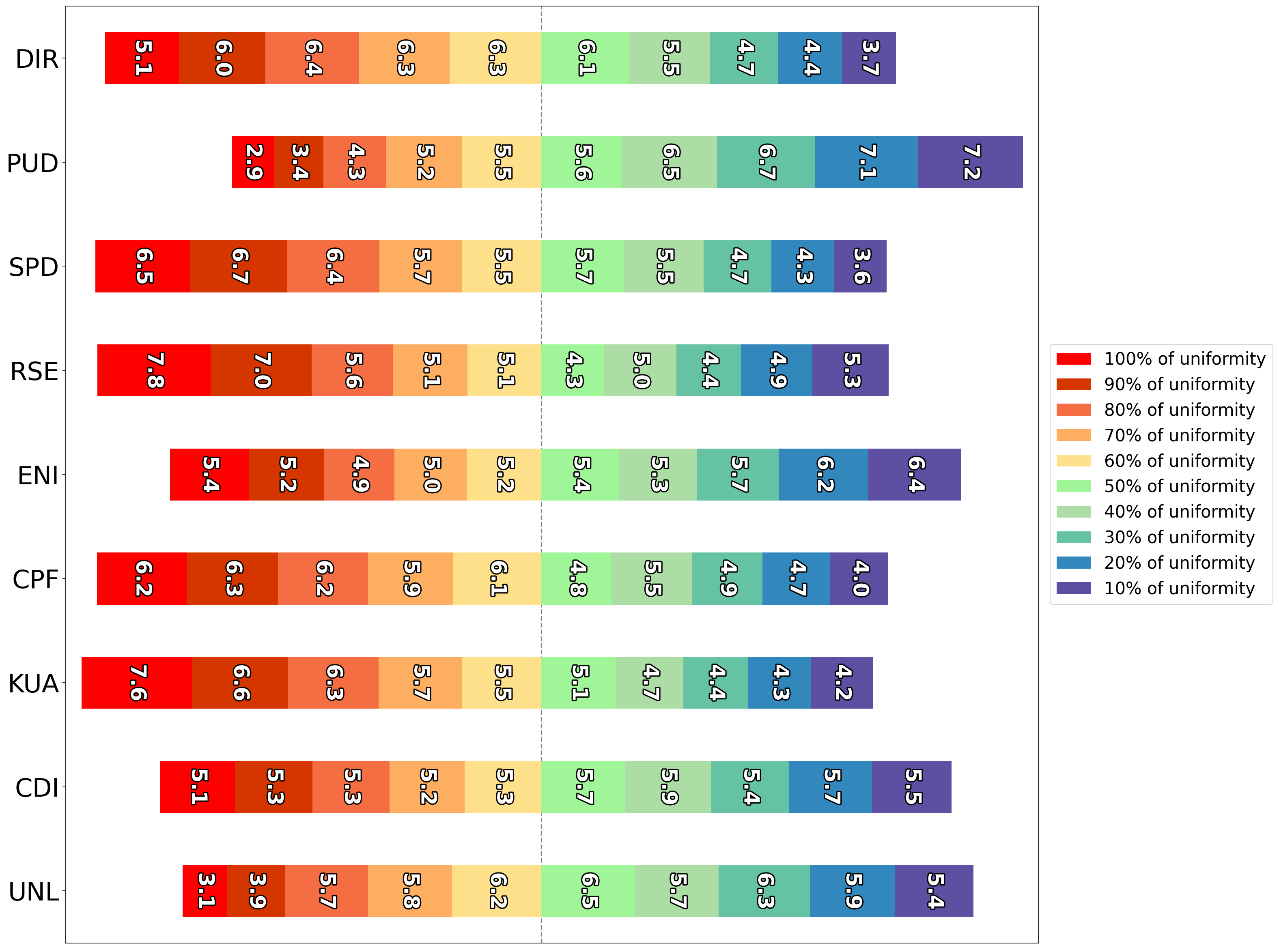}
    \caption{Preferences of the nine DIs for PFAs with different degrees of Uniformity.}
    \label{fig:uniformity_preferences}
\end{figure}

Tables SM-10 to SM-18 in the Supplementary Material contain the numerical values of the DIs assessing all the instances. Figure~\ref{fig:uniformity_preferences} summarizes the numerical results using a Likert Plot with a grading scale from 1 to 10 points, where 10 points are assigned when the DI ranks first the ground truth. The figure shows that RSE assigns a score of 7.8 points to the ground truth. Thus, it prefers PFAs with 100\% of uniformity in most cases. Revising the numerical results in Table SM-13, we can observe that RSE sometimes prefers PFAs with 30\% to 10\% of uniformity, especially when $m\ge 3$. This drives us to think that RSE prefers PFAs with good diversity. The more randomly selected the points, the more diversity they provide in high-dimensional objective spaces. However, it is worth emphasizing that an open problem in EMOO is establishing a mathematical definition of diversity in high-dimensional objective spaces using just hundreds of points to represent the associated manifold. SPD also prefers the ground truth more often than other DIs, but it has problems when assessing ZCAT problems, particularly ZCAT3 and ZCAT4, where it prefers a lower value of uniformity. In contrast, KUA has better preferences on ZCAT instances. Thus, it can be used as a complimentary DI regarding uniformity when no coverage is lost. On the other hand, PUD is biased to prefer less uniform PFAs, thus rewarding PFAs with more diversity. Finally, we should highlight that the ENI, CPF, CDI, DIR, and UNL scores have problems establishing a consistent pattern of preferences, i.e., the PFAs with different degrees of uniformity seem indifferent to them. 

\subsection{Pathological Distributions}
 \input{Tables/Cardinality_pathology}

\begin{figure*}
    \centering
    \begin{subfigure}{0.18\textwidth}
        \includegraphics[width=\textwidth]{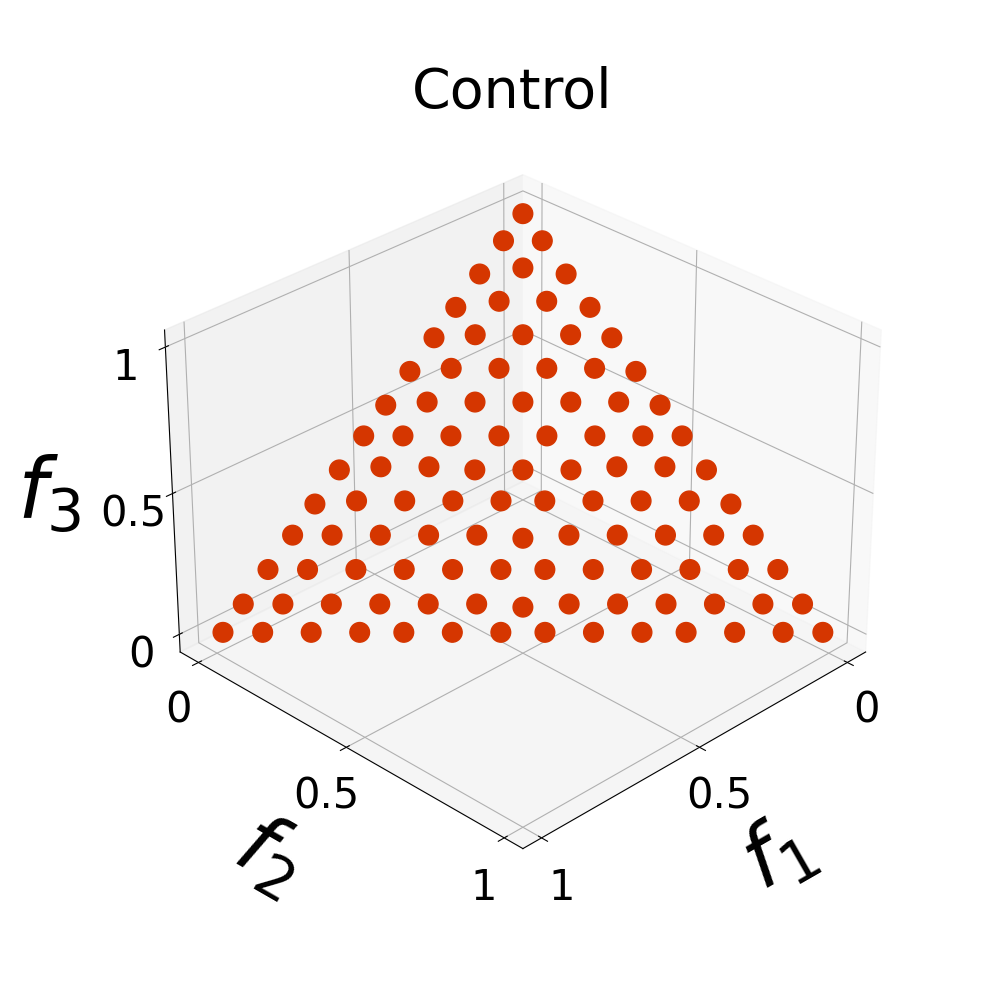}        
    \end{subfigure}
    \hspace{0.2cm}
      \begin{subfigure}{0.18\textwidth}
        \includegraphics[width=\textwidth]{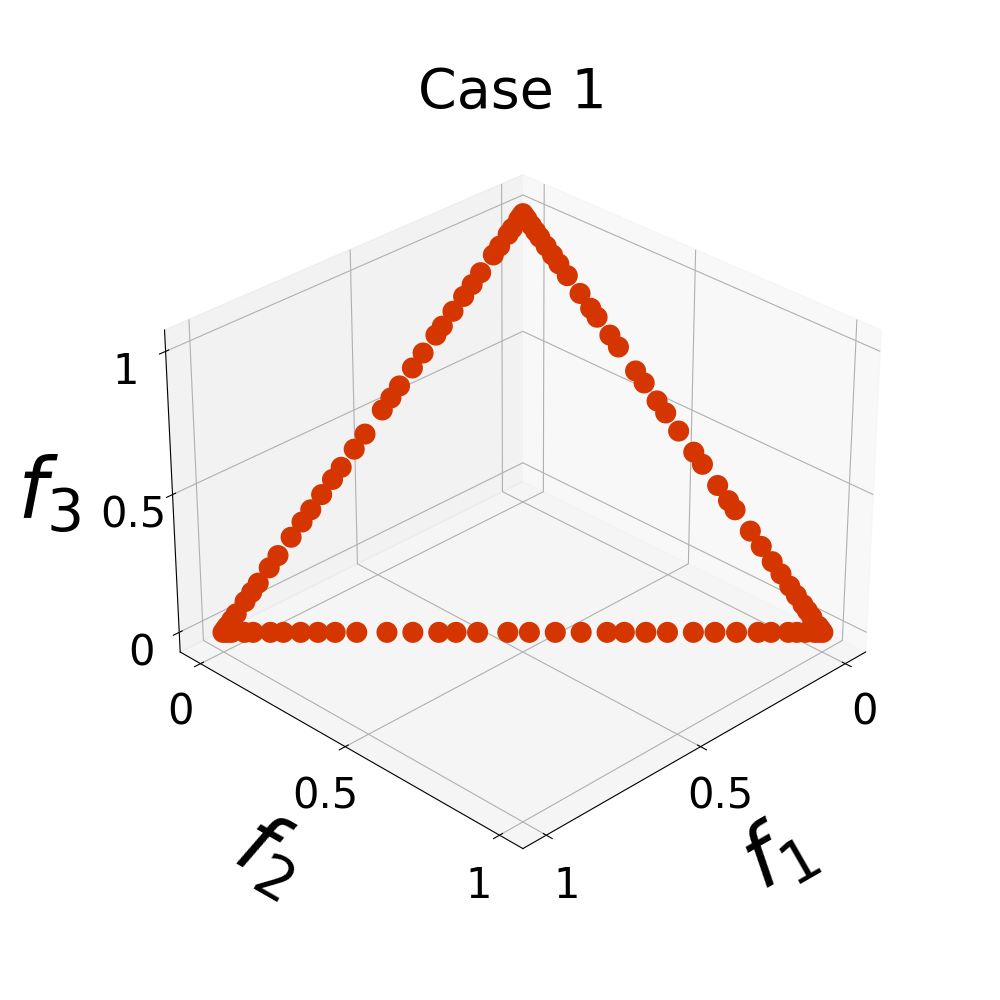}        
    \end{subfigure}
    \hspace{0.2cm}
    \begin{subfigure}{0.18\textwidth}
        \includegraphics[width=\textwidth]{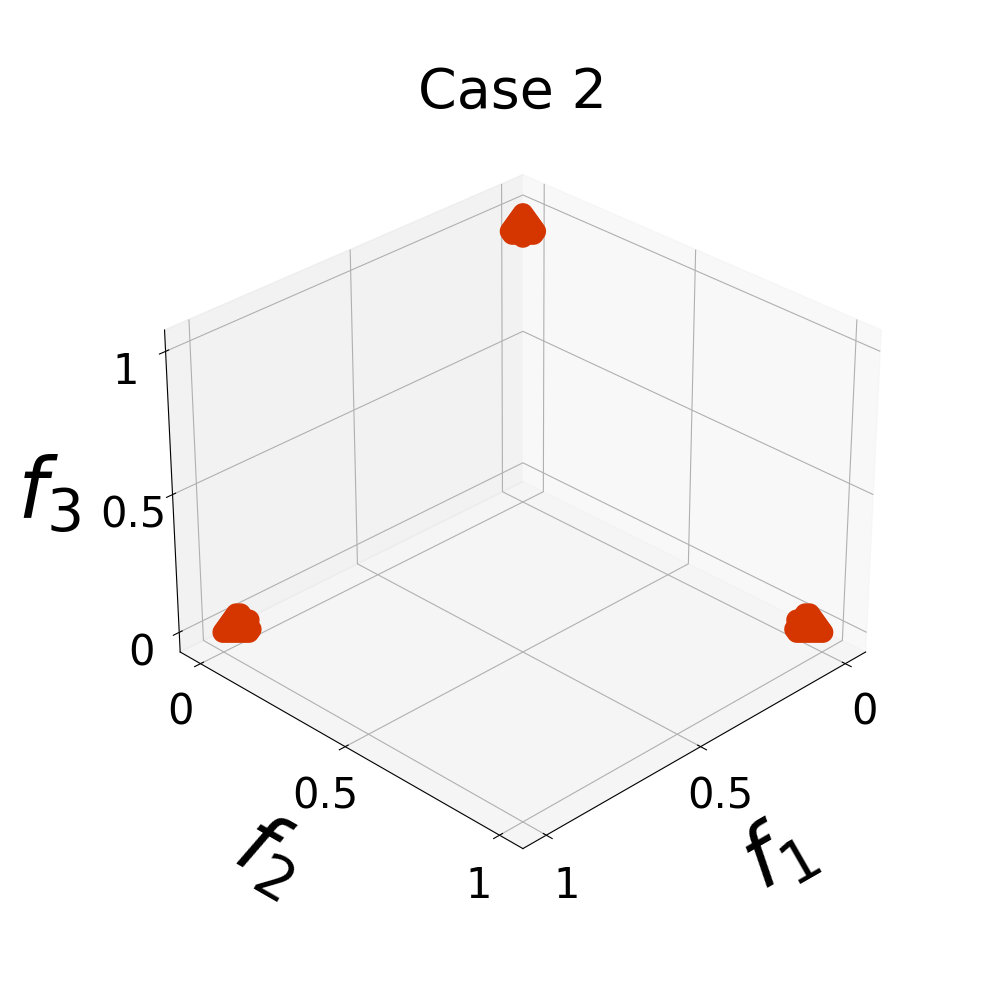}        
    \end{subfigure}
    \hspace{0.2cm}
      \begin{subfigure}{0.18\textwidth}
        \includegraphics[width=\textwidth]{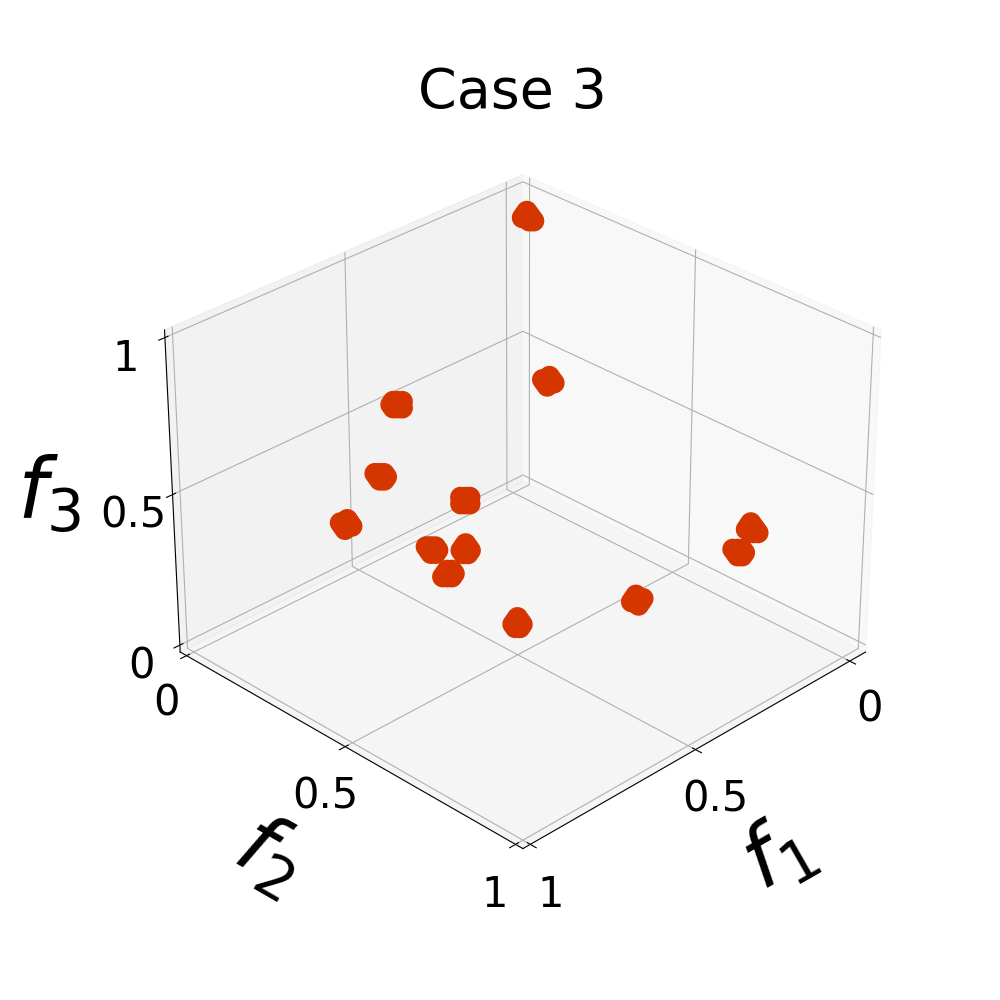}        
    \end{subfigure}



    \caption{Three-objective PFAs for DTLZ1, representing the Control Case and Pathological Distributions. All the PFAs have $N_{100}=105$ points.}\label{fig:pathology}
\end{figure*}

Our third experiment analyzes the preferences of the selected DIs when assessing Pathological Distributions. We defined a Pathological Distribution as a PFA that a bad-performing MOEA may produce. We considered the following three Cases.

\begin{itemize}
    \item \textbf{Case 1: Boundary Solutions}. A MOEA may produce solutions all over the boundary of the Pareto Front with no or a few points in the interior.
    \item \textbf{Case 2: Clustered Extreme Solutions.} A MOEA may produce solutions clusterized around the extreme points of the Pareto Front.
    \item \textbf{Case 3: Clustered Solutions.} A MOEA may produce clustered solutions in random regions of the Pareto Front. 
\end{itemize}

\noindent In addition, we define as a Control Case a PFA with 100\% of Uniformity as in the previous section. Figure~\ref{fig:pathology} compares the three Pathological Distributions and the Control Case for the three-objective DTLZ1. In this analysis, we decided to study the influence of different cardinalities of the PFAs. Thus, we constructed PFAs with cardinalities close to 50, 100, and 200 points, denoted as $N_{50}$, $N_{100}$, and $N_{200}$, as shown in Table~\ref{tab:conf_pathology}. The PFAs of Case 1 were generated using the RSE-based subset selection algorithm proposed in~\cite{Falcon23} with $s=0$. For Case 2, we first identified the extreme points and then used them as centroids for $k$-NN. We followed a similar strategy for Case 3 but generated a random number of centroids.

\begin{figure}
    \centering
    \includegraphics[width=\columnwidth]{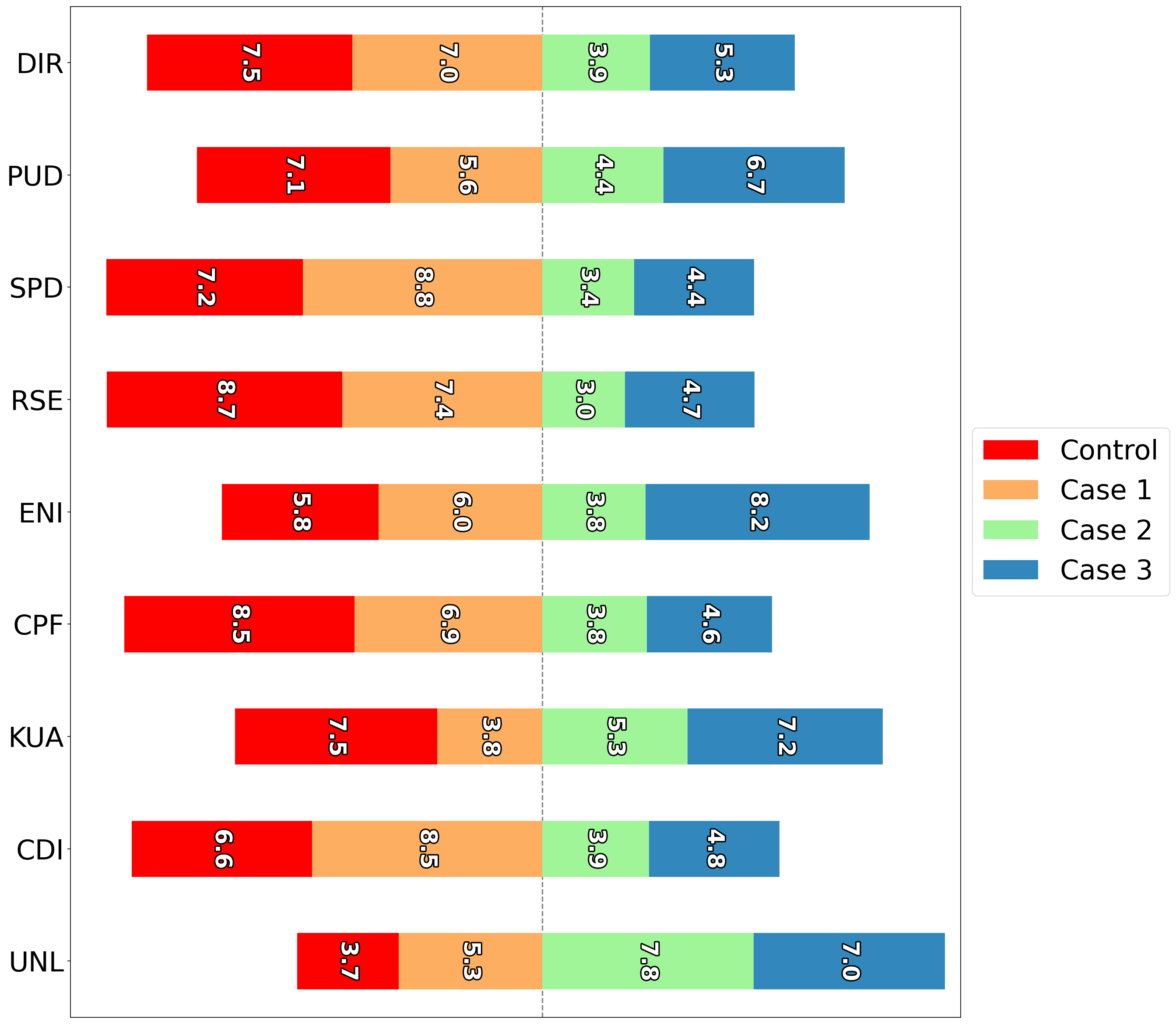}
    \caption{Preferences of the nine DIs for PFAs with $N_{100}$ points for the Control Case and the Pathological Cases 1 to 3.}
    \label{fig:pathology_n100}
\end{figure}

Tables SM-19 to SM-27, SM-28 to SM-36, and SM-37 to SM-45 exhibit the complete numerical results for PFAs with $N_{50}$, $N_{100}$, and $N_{200}$ points, respectively. Figure~\ref{fig:pathology_n100} shows the summary of the results for $N_{100}$ using the Likert Plot. Since we only have four types of PFAs, we used the following grading assignment: 10, 7.5, 5, and 2.5 points are assigned to preference ranks 1, 2, 3, and 4, respectively. As in the previous experiments, the preferences of RSE and SPD have a major tendency to prefer the ground truth, but CPF is also a good option in this case. Regarding the numerical values of RSE in Table SM-31, it does not prefer the Control Case for problems ZCAT1, ZCAT3, and ZCAT in high-dimensional spaces, opting for Case 1. These PFAs of Case 1 may provide more diversity, encouraging RSE to reward them. Regarding the other DIs, only UNL and ENI prefer the Pathological Distributions over the Control Case. 

%% file: Tables/Cardinality_coverage.tex
\begin{table}
    \centering
    \scriptsize
    \caption{Specification of the cardinality of the PFAs as the number of objectives ($m$) increases. $N^*$ is the desired cardinality but $N={\binom{H_1 + m - 1}{m- 1} + \binom{H_2 + m - 1}{m - 1}}$ is obtained using TLSLD with parameters $H_1$ and $H_2$.}
    \label{tab:conf_coverage}
    \begin{tabular}{|c|c|c|c|c|}
        \hline
        {$\mathbf{m}$} & {$\mathbf{N}^*$} & {$\mathbf{N}$} & {$\mathbf{H_1}$} & {$\mathbf{H_2}$} \\
        \hline
        \hline
        2 & 100 & 100 & 99 & 0 \\
        3 & 100 & 105 & 13 & 0 \\
        4 & 150 & 120 & 7 & 0 \\
        5 & 150 & 126 & 5 & 0 \\
        6 & 150 & 147 & 4 & 2 \\
        7 & 200 & 168 & 3 & 3 \\
        8 & 200 & 156 & 3 & 2 \\
        9 & 250 & 210 & 3 & 2 \\
        10 & 250 & 230 & 3 & 1 \\
        \hline
    \end{tabular}    
\end{table}

%% file: Tables/Cardinality_pathology.tex
\begin{table}
    \centering
    \scriptsize
    \caption{Specification of cardinalities $N_{50}$, $N_{100}$, and $N_{200}$ as the number of objectives ($m$) increases. In each case, the number of solutions is equal to $\binom{H_1 + m - 1}{m- 1} + \binom{H_2 + m - 1}{m - 1}$ where $H_1$ and $H_2$.}
    \label{tab:conf_pathology}
    \resizebox{\columnwidth}{!}{
    \begin{tabular}{|c||c|c|c||c|c|c||c|c|c|}
        \hline
         $\mathbf{m}$ & $\mathbf{N_{50}}$ & $\mathbf{H_1}$ & $\mathbf{H_2}$ & $\mathbf{N_{100}}$ & $\mathbf{H_1}$ & $\mathbf{H_2}$ & $\mathbf{N_{200}}$ & $\mathbf{H_1}$ & $\mathbf{H_2}$ \\
         \hline
         \hline
         2 & 50 & 49 & 0 & 100 & 99 & 0 & 200 & 199 & 0 \\
         3 & 45 & 8 & 0 & 105 & 13 & 0 & 210 & 19 & 0 \\
         4 & 56 & 5 & 0 & 120 & 7 & 0 & 220 & 9  & 0 \\
         5 & 50 & 3 & 2 & 105 & 4 & 3 & 210 & 6 & 0 \\
         6 & 42 & 2 & 2 & 112 & 3 & 3 & 258 & 5 & 1 \\
         7 & 56 & 2 & 2 & 91 & 3 & 1 & 210 & 4 & 0 \\
         8 & 44 & 2 & 1 & 120 & 3 & 0 & 240 & 3 & 3 \\
         9 & 45 & 2 & 0 & 90 & 2 & 2 & 210 & 3 & 2 \\
         10 & 55 & 2 & 0 & 110 & 2 & 2 & 220 & 3 & 0 \\
         \hline         
    \end{tabular}
    }
\end{table}

%% file: Sections/S5.tex
\section{Conclusions and Future Work}\label{sec:conclusions}
This study focused on assessing Distribution Indicators (DIs) for evaluating Pareto Front Approximations (PFAs). We introduced a taxonomy to categorize existing DIs in EMOO. Subsequently, we conducted a preference analysis using nine representative DIs from our taxonomy, exploring scenarios such as loss of coverage, uniformity, and pathological distributions. Our experimental findings revealed several insights: 1) All selected DIs are rotationally invariant to PFAs. 2) Certain DIs may yield misleading results and require cautious application for PFA assessment. 3) The Solow-Polasky Diversity Indicator (SPD) and Riesz $s$-energy (RSE) exhibit promise, demonstrating resistance across all considered scenarios, thereby consistently rewarding ground-truth PFAs. Consequently, we suggest a detailed investigation of SPD and RSE, delving into their mathematical properties. Additionally, we advocate for developing new DIs grounded in Biodiversity and Potential Energy concepts.

%% file: main.bbl
\begin{thebibliography}{10}

\bibitem{Brockhoff23}
D.~Brockhoff, M.~Emmerich, B.~Naujoks, and R.~Purshouse, {\em {Many-Criteria
  Optimization and Decision Analysis: State-of-the-Art, Present Challenges, and
  Future Perspectives}}.
\newblock {Natural Computing Series}, Springer, 2023.

\bibitem{Li19}
M.~Li and X.~Yao, ``{Quality Evaluation of Solution Sets in Multiobjective
  Optimisation: A Survey},'' {\em {ACM Computing Surveys}}, vol.~52,
  pp.~26:1--26:38, Mar. 2019.

\bibitem{Audet21}
C.~Audet, J.~Bigeon, D.~Cartier, S.~{Le Digabel}, and L.~Salomon, ``Performance
  indicators in multiobjective optimization,'' {\em {European Journal of
  Operational Research}}, vol.~292, no.~2, pp.~397--422, 2021.

\bibitem{Zitzler00}
E.~Zitzler, K.~Deb, and L.~Thiele, ``{Comparison of Multiobjective Evolutionary
  Algorithms: Empirical Results},'' {\em Evolutionary Computation}, vol.~8,
  no.~2, pp.~173--195, 2000.

\bibitem{Guerreiro21}
A.~P. Guerreiro, C.~M. Fonseca, and L.~Paquete, ``{The Hypervolume Indicator:
  Computational Problems and Algorithms},'' {\em ACM Computing Surveys},
  vol.~54, jul 2021.

\bibitem{Cai22}
X.~Cai, Y.~Xiao, Z.~Li, Q.~Sun, H.~Xu, M.~Li, and H.~Ishibuchi, ``{A
  Kernel-Based Indicator for Multi/Many-Objective Optimization},'' {\em {IEEE
  Transactions on Evolutionary Computation}}, vol.~26, no.~4, pp.~602--615,
  2022.

\bibitem{Emmerich13a}
M.~T. Emmerich, A.~H. Deutz, and J.~W. Kruisselbrink, ``{On Quality Indicators
  for Black-Box Level Set Approximation},'' in {\em EVOLVE - A bridge between
  Probability, Set Oriented Numerics and Evolutionary Computation} (E.~Tantar,
  A.-A. Tantar, P.~Bouvry, P.~D. Moral, P.~Legrand, C.~A. {Coello Coello}, and
  O.~Sch\"{u}tze, eds.), ch.~4, pp.~157--185, Heidelberg, Germany:
  Springer-Verlag. Studies in Computational Intelligence Vol. 447, 2013.
\newblock 978-3-642-32725-4.

\bibitem{Wang17}
H.~{Wang}, Y.~{Jin}, and X.~{Yao}, ``{Diversity Assessment in Many-Objective
  Optimization},'' {\em {IEEE Transactions on Cybernetics}}, vol.~47, no.~6,
  pp.~1510--1522, 2017.

\bibitem{Schott95}
J.~R. Schott, ``Fault {T}olerant {D}esign {U}sing {S}ingle and {M}ulticriteria
  {G}enetic {A}lgorithm {O}ptimization,'' Master's thesis, Department of
  Aeronautics and Astronautics, Massachusetts Institute of Technology,
  Cambridge, Massachusetts, May 1995.

\bibitem{Bandy04}
S.~Bandyopadhyay, S.~Pal, and B.~Aruna, ``{Multiobjective GAs, Quantitative
  Indices, and Pattern Classification},'' {\em {IEEE Transactions on Systems,
  Man, and Cybernetics, Part B (Cybernetics)}}, vol.~34, no.~5, pp.~2088--2099,
  2004.

\bibitem{Meng05}
H.-y. Meng, X.-h. Zhang, and S.-y. Liu, ``"new quality measures for
  multiobjective programming",'' in {\em "Advances in Natural Computation"}
  (L.~Wang, K.~Chen, and Y.~S. Ong, eds.), ("Berlin, Heidelberg"),
  pp.~"1044--1048", "Springer Berlin Heidelberg", 2005.

\bibitem{Sayin00}
S.~Sayın, ``Measuring the quality of discrete representations of efficient
  sets in multiple objective mathematical programming,'' {\em Mathematical
  Programming}, vol.~87, pp.~543--560, May 2000.

\bibitem{Shang21}
K.~Shang, H.~Ishibuchi, and Y.~Nan, ``Distance-based subset selection
  revisited,'' in {\em {Proceedings of the Genetic and Evolutionary Computation
  Conference}}, GECCO '21, (New York, NY, USA), pp.~439--447, Association for
  Computing Machinery, 2021.

\bibitem{Mehr03}
A.~Farhang-Mehr and S.~Azarm, ``{An Information-Theoretic Entropy Metric for
  Assessing Multi-Objective Optimization Solution Set Quality },'' {\em
  {Journal of Mechanical Design}}, vol.~125, pp.~655--663, 01 2004.

\bibitem{Wang12}
K.~Wang, J.~Zheng, and J.~Zou, ``{Neighbor-Distance based Diversity Assessment
  for Multi-Objective Optimizations},'' in {\em {2012 8th International
  Conference on Natural Computation}}, pp.~833--837, 2012.

\bibitem{Li05}
X.-y. Li, J.-h. Zheng, and J.~Xue, ``A diversity metric for multi-objective
  evolutionary algorithms,'' in {\em Advances in Natural Computation} (L.~Wang,
  K.~Chen, and Y.~S. Ong, eds.), (Berlin, Heidelberg), pp.~68--73, Springer
  Berlin Heidelberg, 2005.

\bibitem{Weitzman92}
M.~L. Weitzman, ``{On Diversity$^*$},'' {\em {Quarterly Journal of Economics}},
  vol.~107, no.~2, pp.~363--405, 1992.

\bibitem{Solow94}
A.~R. Solow and S.~Polasky, ``{Measuring biological diversity},'' {\em
  {Environmental and Ecological Statistics}}, vol.~1, no.~2, pp.~95--103, 1994.

\bibitem{Falcon23}
J.~G. Falc{\'o}n-Cardona, E.~{Covantes Osuna}, C.~A. {Coello Coello}, and
  H.~Ishibuchi, ``On the utilization of pair-potential energy functions in
  multi-objective optimization,'' {\em {Swarm and Evolutionary Computation}},
  vol.~79, p.~101308, 2023.

\bibitem{Cai18}
X.~Cai, H.~Sun, and Z.~Fan, ``A diversity indicator based on reference vectors
  for many-objective optimization,'' {\em {Information Sciences}},
  vol.~430--431, pp.~467--486, 2018.

\bibitem{Sergiy19}
S.~V. Borodachov, D.~P. Hardin, and E.~B. Saff, {\em {Discrete Energy on
  Rectifiable Sets}}.
\newblock Springer Monographs in Mathematics, Springer-Verlag, 1~ed., 2019.

\bibitem{Tian19}
Y.~Tian, R.~Cheng, X.~Zhang, M.~Li, and Y.~Jin, ``{Diversity Assessment of
  Multi-Objective Evolutionary Algorithms: Performance Metric and Benchmark
  Problems [Research Frontier]},'' {\em IEEE Computational Intelligence
  Magazine}, vol.~14, no.~3, pp.~61--74, 2019.

\bibitem{Jiang16}
S.~Jiang, S.~Yang, and M.~Li, ``On the use of hypervolume for diversity
  measurement of pareto front approximations,'' in {\em {2016 IEEE Symposium
  Series on Computational Intelligence (SSCI)}}, pp.~1--8, 2016.

\bibitem{Shankar17}
K.~S. Bhattacharjee, H.~K. Singh, T.~Ray, and Q.~Zhang, ``{Decomposition Based
  Evolutionary Algorithm with a Dual Set of Reference Vectors},'' in {\em {2017
  IEEE Congress on Evolutionary Computation (CEC)}}, pp.~105--112, 2017.

\bibitem{Deb05c}
K.~Deb, L.~Thiele, M.~Laumanns, and E.~Zitzler, ``{Scalable Test Problems for
  Evolutionary Multiobjective Optimization},'' in {\em Evolutionary
  Multiobjective Optimization: Theoretical Advances and Applications}
  (A.~Abraham, L.~Jain, and R.~Goldberg, eds.), pp.~105--145, Springer London,
  2005.

\bibitem{Huband06}
S.~Huband, P.~Hingston, L.~Barone, and L.~While, ``{A review of multiobjective
  test problems and a scalable test problem toolkit},'' {\em IEEE Transactions
  on Evolutionary Computation}, vol.~10, no.~5, pp.~477--506, 2006.

\bibitem{Zapotecas23}
S.~Zapotecas-Martínez, C.~A. {Coello Coello}, H.~E. Aguirre, and K.~Tanaka,
  ``Challenging test problems for multi- and many-objective optimization,''
  {\em {Swarm and Evolutionary Computation}}, vol.~81, p.~101350, 2023.

\bibitem{Tian17}
Y.~{Tian}, R.~{Cheng}, X.~{Zhang}, and Y.~{Jin}, ``{PlatEMO: A MATLAB Platform
  for Evolutionary Multi-Objective Optimization [Educational Forum]},'' {\em
  IEEE Computational Intelligence Magazine}, vol.~12, no.~4, pp.~73--87, 2017.

\end{thebibliography}
